\documentclass[10pt,twocolumn,letterpaper]{article}

\usepackage[final]{cvpr}      

\usepackage{graphicx}
\usepackage{amsmath}
\usepackage{amssymb}
\usepackage{booktabs}
\usepackage{cuted}

\usepackage[pagebackref,breaklinks,colorlinks]{hyperref}

\usepackage[capitalize]{cleveref}
\crefname{section}{Sec.}{Secs.}
\Crefname{section}{Section}{Sections}
\Crefname{table}{Table}{Tables}
\crefname{table}{Tab.}{Tabs.}

\begin{document}

\title{Neural Relighting and Expression Transfer on Video Portraits}

\author{
Youjia Wang*\\
{\tt\small wangyj2@shanghaitech.edu.cn}
\and
Taotao Zhou*\\
{\tt\small zhoutt@shanghaitech.edu.cn}
\and
Minzhang Li\\
{\tt\small limzh@shanghaitech.edu.cn}
\and
Teng Xu\\
{\tt\small xuteng@shanghaitech.edu.cn}
\and
Lan Xu\\
{\tt\small xulan1@shanghaitech.edu.cn}
\and
Jingyi Yu\\
{\tt\small jingyiyu@shanghaitech.edu.cn}
\and
ShanghaiTech University\\
}
\maketitle



\begin{abstract}
	Photo-realistic video portrait reenactment benefits virtual production and numerous VR/AR experiences. The task remains challenging as the reenacted expression should match the source while the lighting should be adjustable to new environments.
	We present a neural relighting and expression transfer technique to transfer the head pose and facial expressions from a source performer to a portrait video of a target performer while enabling dynamic relighting. 
	Our approach employs 4D reflectance field learning, model-based facial performance capture and target-aware neural rendering. 
	Specifically, given a short sequence of the target performer's OLAT, we apply a rendering-to-video translation network to first synthesize the OLAT result of new sequences with unseen expressions.
	We then design a semantic-aware facial normalization scheme along with a multi-frame multi-task learning strategy to encode the content, segmentation, and motion flows for reliably inferring the reflectance field. This allows us to simultaneously control facial expression and apply virtual relighting. 
	%
%
%
	%
	%
	Extensive experiments demonstrate that our technique can robustly handle challenging expressions and lighting environments and produce results at a cinematographic quality.
\end{abstract}


\section{Introduction}
The popularity of mobile cameras has witnessed the rapid development of digital facial portrait photography.
Further synthesizing and editing the video portraits suggests different content, which enables numerous applications in virtual cinematography, movie post-production, visual effects and telepresence, among others.
\begin{figure}[t]
  \includegraphics[width=\linewidth]{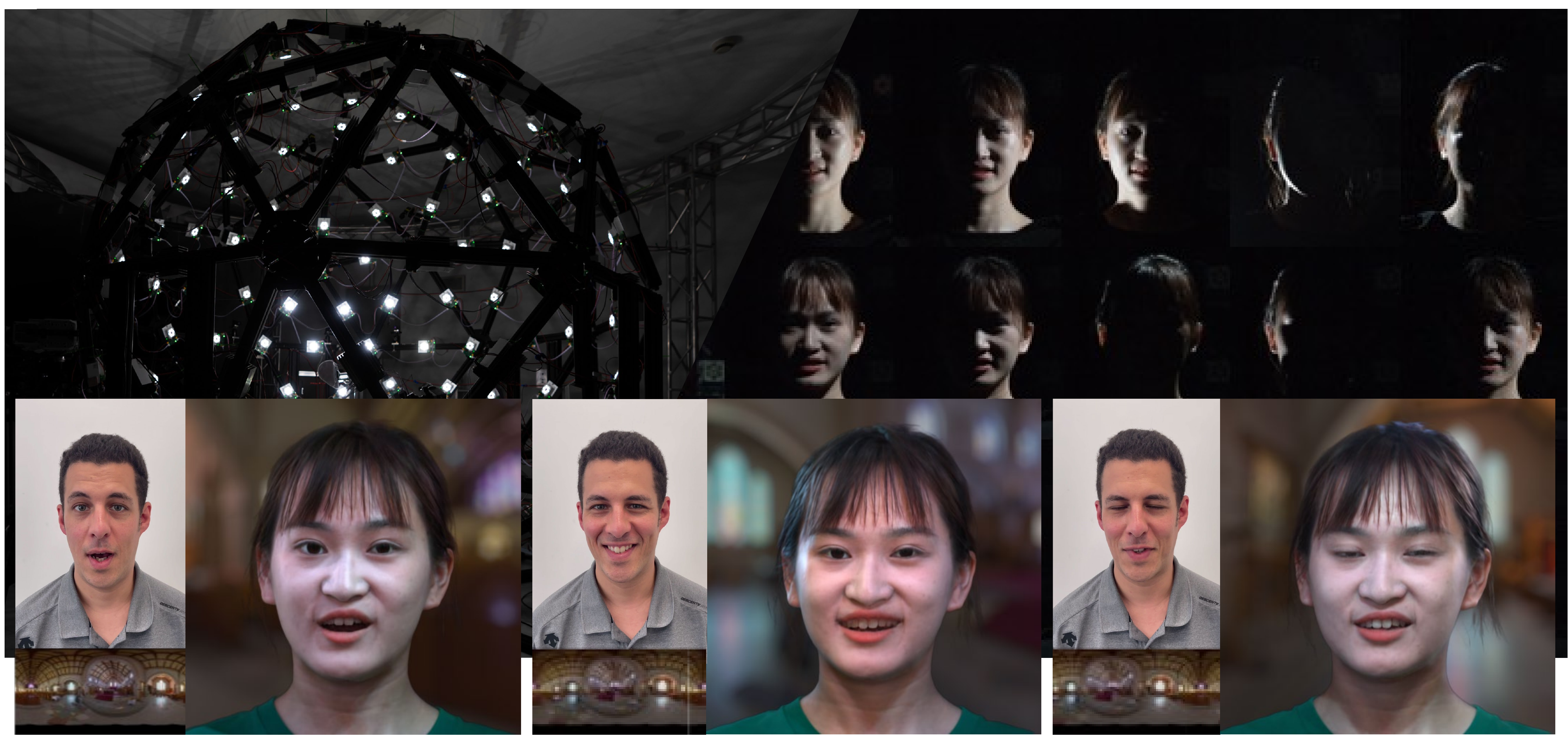}
  \caption{Top: our lighting dome for acquiring the short OLAT sequence of the target performer. Bottom: a different performer can transfer his expressions, unseen from the target sequence, with new lighting conditions at a cinematographic quality.}
  \label{fig:teaser}
\end{figure}
Generating such video portraits conveniently still remains challenging and has attracted substantive attention of both the computer vision and graphics communities.

In this paper, we focus on automatically and conveniently synthesizing a photo-realistic video portrait of a \textit{target} actor where the facial expression, the lighting condition and background are fully disentangled and controllable through various \textit{source} actors.
It is particularly challenging since humans are highly sensitive to facial inconsistency and the video portraits are influenced by the complex interaction of environment, lighting patterns and various facial attributes like material and specular reflection. 
The most common practice for manipulating video portraits in the film industry still relies on tedious manual labor and sophisticated devices for reflectance field acquisition~\cite{debevec2000acquiring} or chroma-keying~\cite{wright2006digital}.
The recent neural techniques~\cite{NR_survey,kim2018deep} bring huge potential for convenient and highly-realistic manipulations, enabling compelling visual quality in various applications, such as visual dubbing~\cite{kim2019neural,ji2021audio}, telepresence~\cite{lombardi2018deep}, video post-editing~\cite{fried2019text,yao2020talkinghead} or audio-driven virtual assistant~\cite{thies2020neural}, even in real-time~\cite{thies2016face2face,thies2018headon}.
However, they still suffer from pre-baked shading of the specific training scenes without manipulating the illumination, which is critical for relighting applications in portrait photography like virtual cinematography.
On the other hand, various neural schemes with controllable portrait's lighting conditions have been proposed using high-quality pore-level 4D facial reflectance fields~\cite{debevec2000acquiring} acquired under multi-view and multi-lit configurations.
However, they are limited to image-based relighting~\cite{sun2019single,pandey2021total}, a subset of the reflectance field learning~\cite{sun2020light,meka2019deep} or relightable model and appearance reconstruction~\cite{guo2019relightables,meka2020deep,bi2021avatar}.
Researchers pay less attention to combine the background and lighting manipulation with face-interior reenactment into a universal data-driven framework.

In this paper, we attack the above challenges and present the first approach to generating relightable neural video portraits for a specific performer. 
As illustrated in Fig.~\ref{fig:teaser}, with the aid of immediate and high-quality facial reflectance field inference, our approach endows the entire photo-realistic video portraits synthesis with the ability of fully disentanglement and control of various head poses, facial expressions, backgrounds, and lighting conditions.
This unique capability also enables numerous photo-realistic visual editing effects.

Our key idea is to combine high-quality 4D reflectance field learning, model-based facial performance capture and target-aware neural rendering into a consistent framework, so as to support convenient, realistic and controllable editing of neural video portraits. 
To this end, for the temporal ground-truth supervision, we first build up a light stage setup to collect the dynamic facial reflectance fields of the target performer with one-light-at-a-time (OLAT) images under various sequential poses and expressions.
%
%
Then, with such target-specific OLAT data, the core of our approach is a novel rendering-to-video translation network design with the explicit head pose, facial expression, lighting and background disentanglement for highly controllable neural video portrait synthesis.
After training, reliable head pose and facial expression editing are obtained by applying the same facial capture and normalization scheme to the source video input, while our explicit OLAT output enables high-quality relit effect. This relit effect can not only match the portrait with the environment, but also simulate the rim-light to improve the matting quality for a better background replacement.
An encoder-decoder architecture is further adopted to encode the illumination weighting parameters of OLAT images from the corresponding relit images, enabling automatic lighting editing from natural source images.
Thus, our approach enables automatically and conveniently control of facial expression, lighting condition and background of the performer's video portrait with high photo-realism for various visual editing effects.
To summarize, our main contributions include: 
\begin{itemize} 
	\setlength\itemsep{0em}
	\item We demonstrate the new capability of simultaneous relighting and and expression transfer, which enables photo-realistic video portraits synthesis for various visual editing effects, with full disentanglement of expressions, backgrounds and lighting conditions.
	
	\item We introduce a rendering-to-video translation network to transfer model-based input into high-fidelity facial reflectance fields, with a multi-frame multi-task learning strategy to encode content, segmentation and temporal information.
		
	\item We propose to utilize hybrid model-based facial capture with a carefully designed semantic-aware facial normalization scheme for reliable disentanglement.
	 
\end{itemize} 
\begin{figure*}[t]
	\begin{center}
		\includegraphics[width=0.95\linewidth]{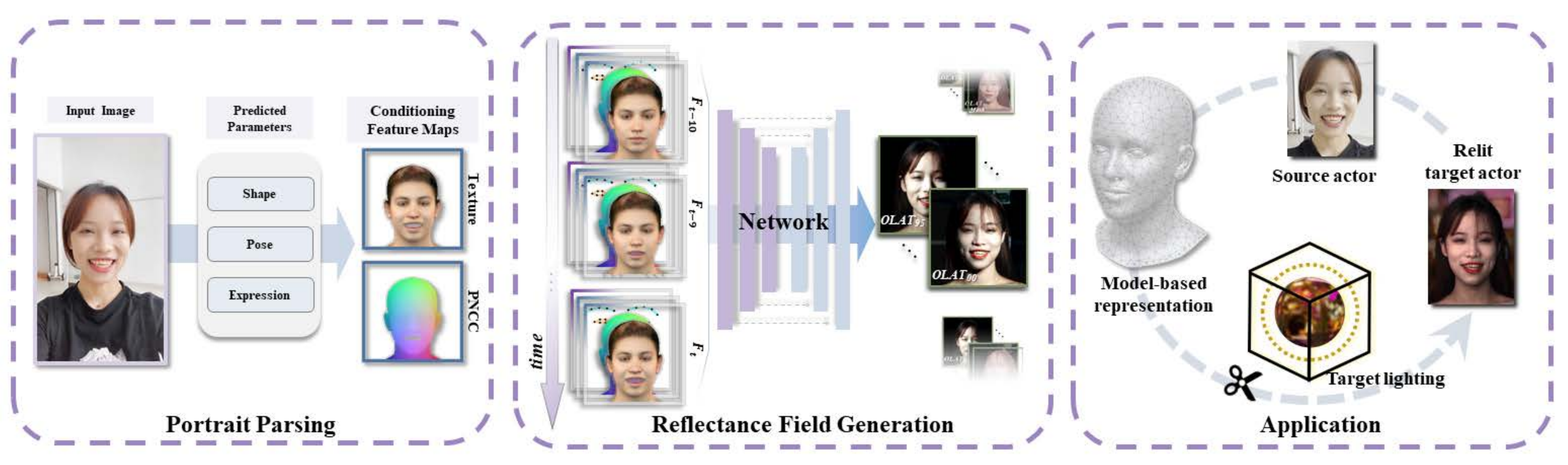}
	\end{center}
	\vspace{-0.5cm}
	\caption{The algorithm pipeline of our relightable neural video portrait. Our pipeline mainly consists of three stages: Portrait parsing disentangles portrait parameters explicitly and normalizes them into the target actor's distribution for conditioning feature map generation; Then the proposed network synthesizes 4D reflectance fields via sequential conditioning feature maps; Our approach can achieve simultaneous relighting and reenactment effects on the portrait video of the target actor.}
	\label{fig:overview}
	\vspace{-0.3cm}
\end{figure*}

\section{Related Work}
\textbf{Face capture and reconstruction.}
 Face reconstruction methods aim to reconstruct the geometry and appearance of 3D face models from visual data and we refer to two recent reports on monocular 3D face reconstruction and applications~\cite{zollhofer2018state,egger20203d} for a comprehensive overview.
Early solutions adopt optimization schemes by fitting a 3D template model of only the face regions into various visual input, such as a single image~\cite{blanz2004exchanging,blanz1999morphable}, a temporal consistent video~\cite{cao2014displaced,fyffe2014driving,suwajanakorn2014total,ichim2015dynamic,garrido2016reconstruction,wu2016anatomically} or even unstructured image sets~\cite{kemelmacher2010being,kemelmacher2013internet,roth2016adaptive}.
The recent deep learning techniques bring huge potential for face modeling, which learn to predict the 3D face shape, geometry or appearance~\cite{kazemi2014one,zhu2017face,tewari2017mofa,tewari2018self,tewari2019fml,cao2015real,richardson20163d,richardson2017learning,yamaguchi2018high}
%
However, these method above still cannot create a photo-realistic model in a controllable manner, especially for those fine-grained facial regions like hair, mouth interior or eye gaze.
%
The recent work~\cite{gafni2020dynamic} enables controllable implicit facial modeling and rendering using neural radiance field,  but it still relies on per-scene training and suffers from pre-baked illumination of the training scenes.
In contrast, our approach enables more disentanglement and control of background, lighting condition as well as facial capture parameters for photo-realistic video portrait generation.

\textbf{Face reenactment and replacement.}
Facial reenactment re-generates the face content of a target actor in a portrait video by transferring facial expression and pose from a source actor.
Many recent reenactment approaches model-based expression capturing and the expressions are transferred via dense motion fields~\cite{liu2011realistic,suwajanakorn2015makes,averbuch2017bringing} or facial parameters~\cite{vlasic2006face,garrido2014automatic,li2013data,thies2015real,thies2016face2face,thies2018facevr}.
The recent neural approaches~\cite{NR_survey,kim2018deep} replace components of the standard graphics pipeline by learned components, which bring huge potential for convenient and highly-realistic manipulations, enabling compelling visual quality in various applications, such as visual dubbing~\cite{kim2019neural,ji2021audio}, telepresence~\cite{lombardi2018deep}, video post-editing~\cite{fried2019text,yao2020talkinghead} or audio-driven virtual assistant~\cite{thies2020neural}, even in real-time~\cite{thies2016face2face,thies2018headon}.
Inspired by pix2pix~\cite{isola2017image}, Deep Video Portraits~\cite{kim2018deep} proposes an rendering-to-image translation approach that converts synthetic rendering input to a scene-specific realistic image output. 
%
%
%
%
%
%
%
However, ~\cite{kim2018deep} and the work followed by, such as ~\cite{kim2019neural, ji2021audio, thies2019deferred}  suffers from pre-baked shading of the specific training scenes without manipulating the illumination which is critical for relighting applications like virtual cinematography.
Some recent work enables lighting or facial expression control during the facial portrait image generation~\cite{tewari2020stylerig,chen2020free} or editing~\cite{tewari2020pie,tewari2021photoapp} based on styleGAN~\cite{karras2019style}.
However, these methods mainly focus on the task of portrait image generation and suffers from unrealistic interpolation for neural video portrait reenactment.
Differently, we take advantage of learning from the temporally consistent OLAT dataset of the target performer to allow for larger changes in facial expression reenactment, background as well as the lighting condition, while maintaining high photo-realism.
\textbf{Facial portrait relighting.}
High quality facial portrait relighting requires the modeling of pore-level 4D reflectance fields.
Debevec~\emph{et al.}~\cite{debevec2000acquiring} invent Light Stage to capture the reflectance field of human faces, which has
enabled high-quality 3D face reconstruction and illuminations rendering, advancing the film’s special effects industry. 
Some subsequent work has also achieved excellent results by introducing deep learning~\cite{zhang2021neural,meka2019deep,meka2020deep,guo2019relightables,sun2019single}.
Some work follows the pipeline of color transfer to achieve the relighting effects~\cite{chen2011face,shih2014style,shu2017portrait,song2017stylizing}, which usually needs another portrait image as the facial color distribution reference. 
With the advent of deep neural networks and neural rendering, some methods~\cite{zhou2019deep,liu2021relighting,sengupta2018sfsnet} adopt Spherical Harmonics (SH) lighting model to manipulate the illumination. 
Several works~\cite{aldrian2012inverse,egger2018occlusion, wang2007face} jointly estimate the 3D face and SH [5, 34] parameters and achieved relighting by recovering the facial geometry and modify the parameters of the SH lighting model. 
Explicitly modeling the shadow and specular~\cite{wang2020single,nestmeyer2020learning} achieve excellent results in directional light source relighting. 
Mallikarjunr~\emph{et al.}~\cite{tewari2020monocular} take a single image portrait as input to predict OLAT(one-light-at-a-time) as Reflectance Fields, which can be relit to other lighting via image-based rendering. 
Sun~\emph{et al.}~\cite{sun2019single} choose environment map as lighting model and use light stage captured OLAT data to generate realistic training data and train relighting networks in an end-to-end fashion. 
Similarly, we use the target-aware temporal OLAT images to generate training data for high-quality lighting disentanglement.
Differently, we further combine the background and lighting manipulation with face-interior reenactment into a universal data-driven framework.
\begin{figure}[t]
  \includegraphics[width=\linewidth]{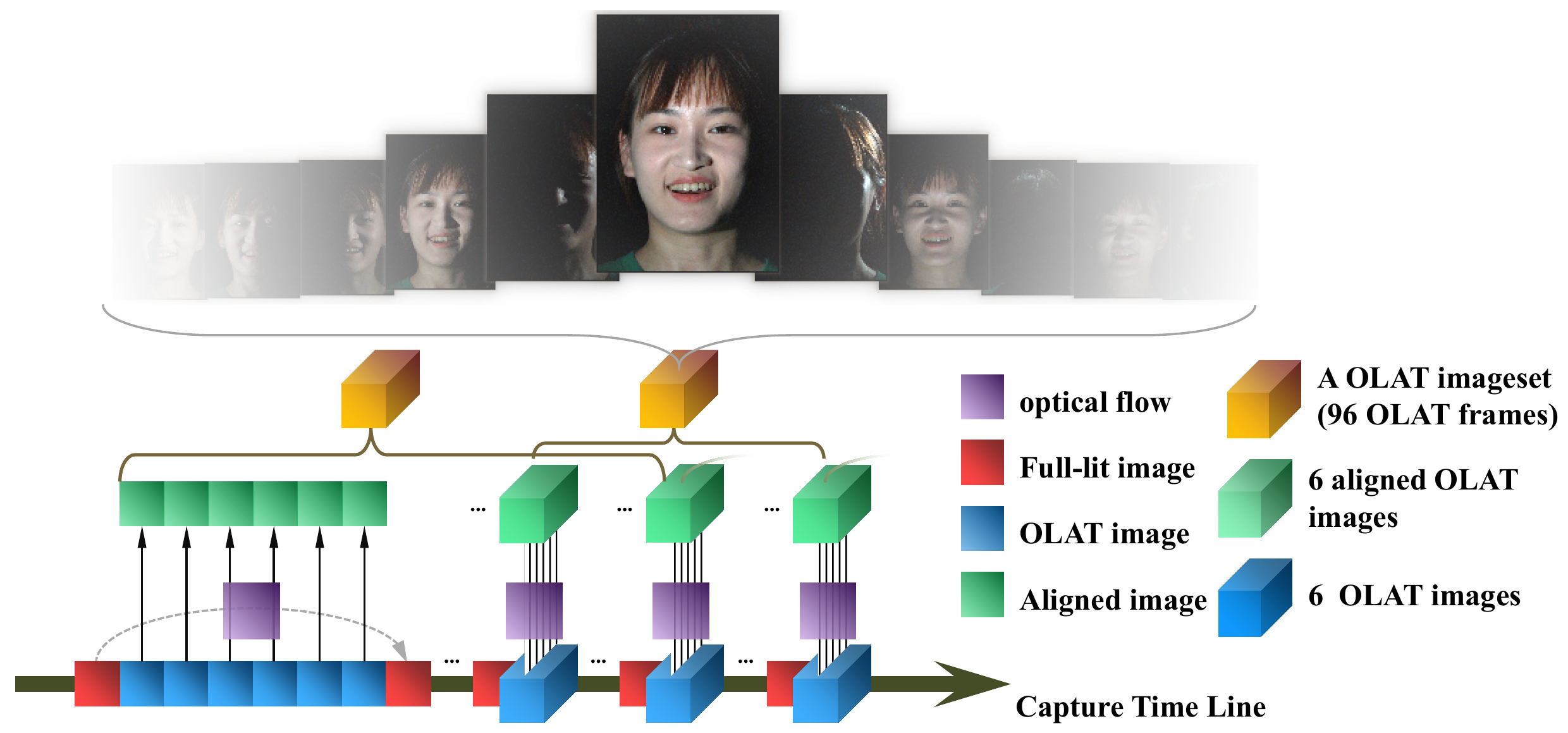}
  \caption{The illustration of capture timing and frame alignment. our method computes optical flows between full-lit images and warps OLAT images accordingly. The "overlapping" strategy allow us to reuse a same OLAT image in different OLAT imageset so that we can achieve higher capture frame rate.}
  \label{fig:optical_flow} 
\end{figure}

\section{Neural Light and Expression Transfer on Video Portraits}\label{sec:method}

We present relightable neural video portrait, a simultaneous relighting and reenactment scheme that transfers facial expressions from a source actor to the portrait video of a target actor with arbitrary new backgrounds and lighting conditions. 
Fig.~\ref{fig:overview} illustrates our technical pipeline, which combines 4D reflectance field learning, model-based facial performance capture and target-aware neural rendering. 
We first describe our dynamic OLAT dataset capture scheme in Sec.~\ref{sec:capture}, 
and further introduce a portrait parsing scheme in Sec.~\ref{sec:parsing} to generate hybrid conditioning feature maps for fully controllable portrait video synthesis.
Specifically, our approach parses the input video of the source actor for the explicit rigid head pose, and facial expression, to support 
explicit editing and disentanglement.
%
%
In sec.~\ref{sec:OLAT_Generatation}, we introduce our rendering-to-video translation network to synthesize high-quality OLAT imagesets from sequential conditioning feature maps. 
We adopt a multi-frame multi-task learning strategy to encode content, segmentation and temporal information simultaneously for reflectance field inference.
Finally, various novel visual editing applications are as described in Sec.~\ref{sec:applications}.

\subsection{Data Acquisition}\label{sec:capture}
To recover the 4D reflectance fields of dynamic portrait, we build up a light stage, so as to provide fine-grained facial perception.
Our hardware architecture is demonstrated in Fig.~\ref{fig:capture_system}, which is a spherical dome of a radius of 1.3 meters with 96 fully programmable LEDs and a 4K ultra-high-speed PCC (Phantom Capture Camera) as shown in Fig.~\ref{fig:capture_system}.

To densely acquire dynamic sets of facial expressions, captured targets were required to perform natural conversations and a range-of-motion sequence with translation and rotation. 

%
However, one of the most challenging issue is that the motion of the captured target will cause mis-alignments, leading to blurriness.
We conquer such limitations by optical flow algorithm and further obtain results at higher frame rate.

%
Inspired by the approach~\cite{meka2020deep}, we additionally capture an full-bright image for tracking purposes every 6 images, as shown in Fig.~\ref{fig:optical_flow}. Then, we align the OLAT data between 14 consecutive groups of full-bright frames with optical flow.
We will discuss the capture setting in supplementary materials for details.
\begin{figure}[t]
  \includegraphics[width=\linewidth]{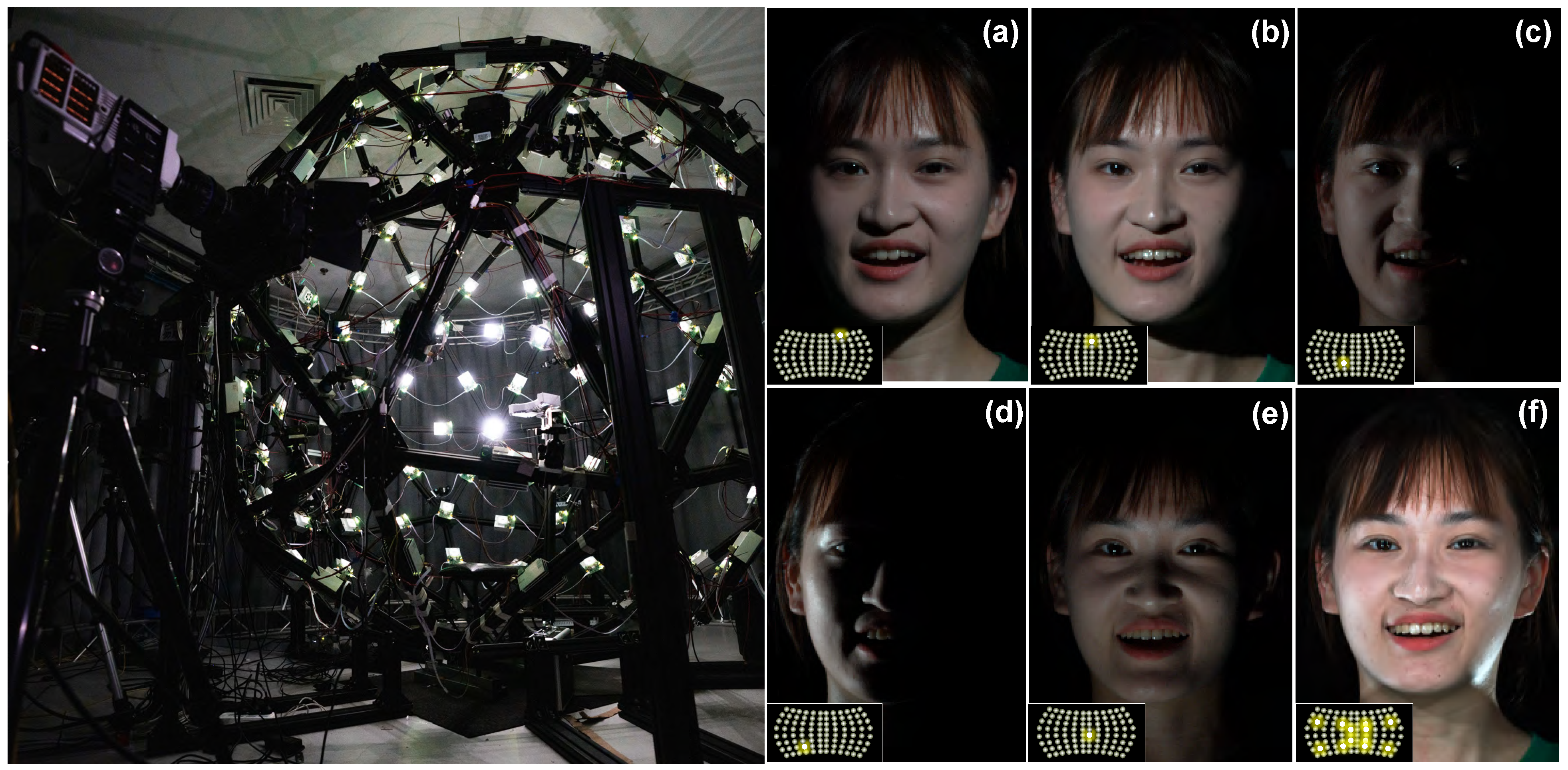}
  \caption{The demonstration of our capturing system and samples of captured data. Left: cameras and lights are arranged on a spherical structure; (a)-(e): samples of OLAT images; (f) a full-lit frame image. Subfigures on the bottom-left of (a)-(f) illustrates the corresponding lighting conditions on our system.}
  \label{fig:capture_system}
\end{figure}

\subsection{Portrait Parsing}\label{sec:parsing}
Our Portrait parsing module aims to generate a set of conditioning feature map $\{\mathbf{F}_t\}_{t=1}^{n_t}$ as inputs to our rendering-to-video network, where $n_t$ is the number of the frames.

A conditioning feature maps $\mathbf{F}_t$ consists of a diffuse color image $\mathbf{I}_t^{d}$ and a coordinate image $\mathbf{I}_t^{c}$. We use FLAME~\cite{li2017learning}, a parametric 3D head model described by a function $G(\mathbf{\beta}, \mathbf{\theta}, \mathbf{\phi})$ as our representation of portrait, where the coefficient parameters $\mathbf{\beta}$, $\mathbf{\theta}$, and $\mathbf{\phi}$ represent head shape, head pose and facial expression, respectively. 
Inspired by DVP~\cite{kim2018deep}, we use diffuse color and pncc as textures for rendering, and use them as the coordinate images $\mathbf{I}_t^{d}$ and $\mathbf{I}_t^c$ respectively.
%

%
%
These images provide a coarse estimation of the detailed facial geometry and spatial information for result synthesis.
%
%
%



Face parameters $\mathbf{\beta}$, $\mathbf{\theta}$, and $\mathbf{\phi}$ are extracted from this image patch via the DECA method~\cite{feng2020learning}, while the 2D landmarks are estimated using a regression trees approach~\cite{kazemi2014one}.

Note that when the source and target actors are different, their facial geometry will be different even in similar facial expressions, which will manifest in facial characteristics, such as eye size, nose length, mouth curvature, etc. Therefore, we performed normalization operations during the expression transfer process. For details, please refer to the supplementary material.

\subsection{Reflectance Field Generation}\label{sec:OLAT_Generatation}
\begin{figure*}[t]
    \centering
	\includegraphics[width=1.0\linewidth]{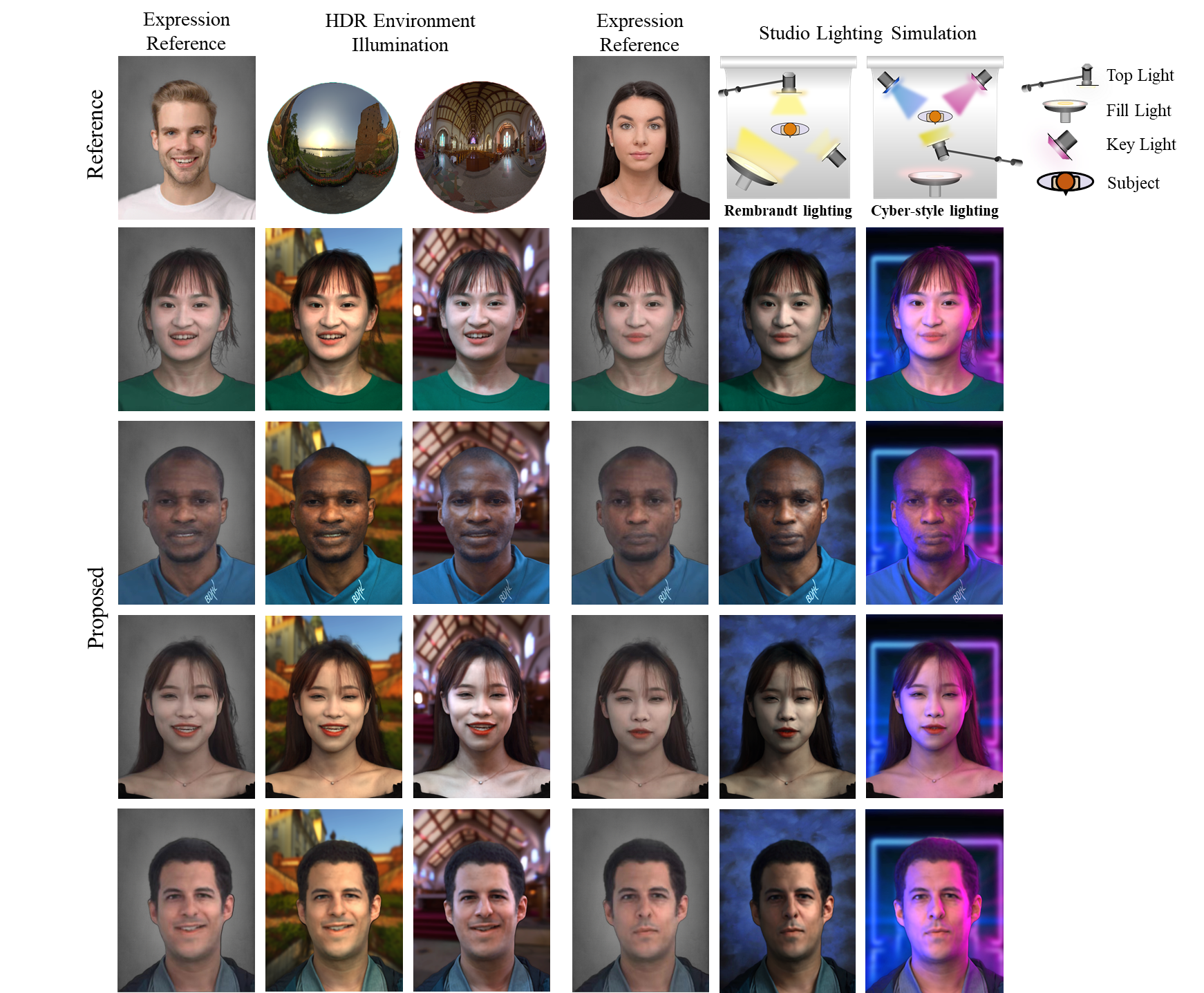}
	\caption{\textbf{Relighting and expression transfer.} The first row is the reference for expression and the reference for lighting. Each following row shows the result of a target actor under the reference expression and lighting. For columns 1 and 4, we use an ambient light to relit, to show our expression transfer result. For each expression, we present two additional lighting condition. In columns 2 and 3, we use HDR environment illumination to relit the target actor, and use the corresponding region in the environment map as the background. In columns 5 and 6, we show our neural reflectance field's simulation of studio lighting, and use a suitable picture as the background.}
	\label{fig:gallery}
\end{figure*}
Our conditioning rendering-to-video translation network takes the composed sequence of conditioning feature maps as input.
Here, the generated reflectance field consists of $96$ portrait images in the same head pose and facial expression from an identical camera view but are lit one light at a time. 
Such design in our approach enables more controllable high-quality relighting than directly predicting the portrait image under specific environment lighting conditions. 
Our explicit relighting strategy based on OLAT imagesets owns better generalization ability than those end-to-end learning ones which are highly relied on the diversity of training data. 
Meanwhile, our target-aware neural rendering strategy can render not only photo-realistic facial output even for those regions complying with head motion such as hair. 
As illustrated in Fig.~\ref{fig:overview}, our translation network adopts a multi-frame multi-task learning framework, which will be described in detail below.

\textbf{Sequential input.}
We gather the adjacent $11$ frames to be a sequential input $\mathcal{F}_{t_c}= \{\mathbf{F}_t\}_{t=t_c-10}^{t_c}$ for the network inference, where $t_c$ is the current time. 
This sequential input enables the network to extract the rich temporal information in our dynamic OLAT dataset and output stabilized image sequence. 
%
 

\textbf{Translation network.}
%
%
%
%
The encoder extracts multi-scale latent representations of conditioning feature maps, while the decoder module generate the reflectance field ,as shown in Fig.~\ref{fig:overview}.
Many works have shown that learning geometric information while learning reflectance helps the neural network to disambiguate the relationship between light and albedo. Therefore, we design the network to explicitly learn the information of albedo and normal. Such a multi-task framework enforces the encoder and the decoder to learn contextual information of the target actor, so as to produce more detailed reflectance fields. Please refer to supplementary for more details

\textbf{Training details.}
We leverage captured dynamic OLAT sequences of the target actor to train our target-aware translation network. 
%
%
We synthesize a fully-illuminated image via relighting the corresponding OLAT imageset $\mathbf{Y}_t$ for each frame. 
Portrait parsing will compute $\{\beta_t\}_{t=1}^{n_t}$, $\{\theta_t\}_{t=1}^{n_t}$, $\{\phi_t\}_{t=1}^{n_t}$, and $\{\mathbf{\hat{I}}_t^l\}_{t=1}^{n_t}$ and further render and compose them into conditioning feature maps $\{\mathbf{F}_t\}_{t=1}^{n_t}$ as described in Section~\ref{sec:parsing}. 

%
%
So we compute the average head shape parameter $\mathbf{\bar{\beta}}= \frac{\sum_{t=1}^{n_t}\mathbf{\beta}_t}{n_t}$ representing the actor's head shape which will be used in synthesizing conditioning inputs. 

We utilize Mean Square Error(MSE) loss and the Multi-scale Structural Similarity(MS-SSIM) Index~\cite{1284395} loss to penalize the differences between the synthesized reflectance field and the ground truth, which are combined together as a photometric loss $\mathcal{L}_{color}$:
\begin{equation}
\begin{aligned}
\mathcal{L}_{color}(\mathbf{\alpha}) = \mathbb{E}_{\mathcal{F},\mathbf{Y}}[\|\mathbf{Y}-\Psi_o(\mathcal{F}; \mathbf{\alpha})\|_2^2 + f_{mss}(\mathbf{Y},\Psi_o(\mathcal{F}; \mathbf{\alpha}))], 
\end{aligned}
\label{eq:photometric}
\end{equation}
where $f_{mss}(\cdot)$ is a differentiable MS-SSIM function~\cite{1284395}; $\mathbf{\alpha}$ is the network weights of the proposed translation network; $\Psi_o(\cdot)$ outputs the predicted reflectance field of the network. 
%
%

We designed loss for the two structural information of albedo and normal:
\begin{equation}
\begin{aligned}
\mathcal{L}_{normal}(\mathbf{\alpha}) = &\mathbb{E}_{\mathbf{N}}[1 -  f_{angle}(\mathbf{N}, \Psi_n(\mathbf{\alpha}))],\\
\mathcal{L}_{albedo}(\mathbf{\alpha}) = &\mathbb{E}_{\mathbf{A}}[ f_{vgg}(\mathbf{A}, \Psi_a(\mathbf{\alpha}))+f_{mss}(\mathbf{A}, \Psi_a(\mathbf{\alpha}))],
\end{aligned}
\label{eq:normal}
\end{equation}
where $f_{angle}(\cdot)$ denotes the per-pixel cosine distance between the two vector ~\cite{Abrevaya_2020_CVPR}, $f_{vgg}(\cdot)$ denotes the output feature of the third-layer of pretrained VGG-19; $\Psi_n(\cdot)$, $\Psi_a(\cdot)$ outputs the predicted normal map, predicted albedo map of the network respectively.

Meanwhile, we deploy an adversarial loss $\mathcal{L}_{GAN}$ to enhance the similarity of the distribution between predicted reflectance field and the ground truth so that the proposed network can produce photo-realistic results. 
Our discriminator $D(\cdot)$ is inspired by the PatchGAN~\cite{DBLP2018patch} classifier and has a similar architecture but difference in inputs. $D(\cdot)$ conditions on the input, the conditioning feature maps $\mathcal{F}$, and either the predicted $\Psi_o(\mathcal{F}; \mathbf{\alpha})$ or the ground-truth reflectance field $\mathbf{Y}$. 
The adversarial loss $\mathcal{L}_{GAN}$ has the following form:
\begin{equation}
\begin{aligned}
\mathcal{L}_{GAN}(\mathbf{\alpha},\mathbf{\omega}) =  &\mathbb{E}_{\mathcal{F},\mathbf{Y}}[\log D(\mathcal{F},\mathbf{Y};\mathbf{\omega})] \\
+ &\mathbb{E}_{\mathcal{F}}[\log(1-D(\mathcal{F}, \Psi_o(\mathcal{F}; \mathbf{\alpha}); \mathbf{\omega}))],
\end{aligned}
\label{eq:gan_loss}
\end{equation}
where $\mathbf{\omega}$ is the network weights of the discriminator $D(\cdot)$.

The color reflectance field branch directly outputs a reflectance field with $96$ portrait images.
The lack of constraints on these images let the network's outputs have slight facial geometry differences, which is not expected in our task and will affect the relighting quality. 
We propose a context loss to alleviate this inconsistency in the predicted reflectance field. 
Specifically, we re-render a fully-illuminated portrait image from the predicted reflectance field which is a differentiable linear combination of output OLAT images. 
Next, we extract the head shape parameters $\mathbf{\beta}(\Psi_o(\mathcal{F}; \mathbf{\alpha}))$ of the re-rendered portrait image and expect it to be as close to $\mathbf{\bar{\beta}}$ as possible. 
The context loss $\mathcal{L}_{context}$ is formulated as:
\begin{equation}
\begin{aligned}
\mathcal{L}_{context}(\mathbf{\alpha}) = \mathbb{E}_{\mathcal{F}}(\|\mathbf{\beta}(\Psi_o(\mathcal{F}; \mathbf{\alpha}))-\mathbf{\bar{\beta}}\|_1).
\end{aligned}
\label{eq:context_loss}
\end{equation}

We linearly combine these loss function as the following objective function to find the optimal weights of translation network:
\begin{equation}
\begin{aligned}
\mathbf{\alpha}^*  &= \mathop{\arg\min}_{\mathbf{\alpha}}\mathop{\max}_{\mathbf{\omega}}   \lambda_1\mathcal{L}_{GAN}(\mathbf{\alpha},\mathbf{\omega}) + \mathcal{L}_{color}(\mathbf{\alpha})\\
& + \mathcal{L}_{normal}(\mathbf{\alpha}) + \mathcal{L}_{albedo}(\mathbf{\alpha}) + \lambda_2\mathcal{L}_{context}(\mathbf{\alpha}),
\end{aligned}
\label{eq:photometric2}
\end{equation}
where $\lambda_1$ and $\lambda_2$ are weights to balance the contribution of these terms. We set $\lambda_1=\lambda_2= 0.1$ in our implementation.

%

\subsection{Applications} \label{sec:applications}

\begin{figure}[t]\centering
  \includegraphics[width=\linewidth]{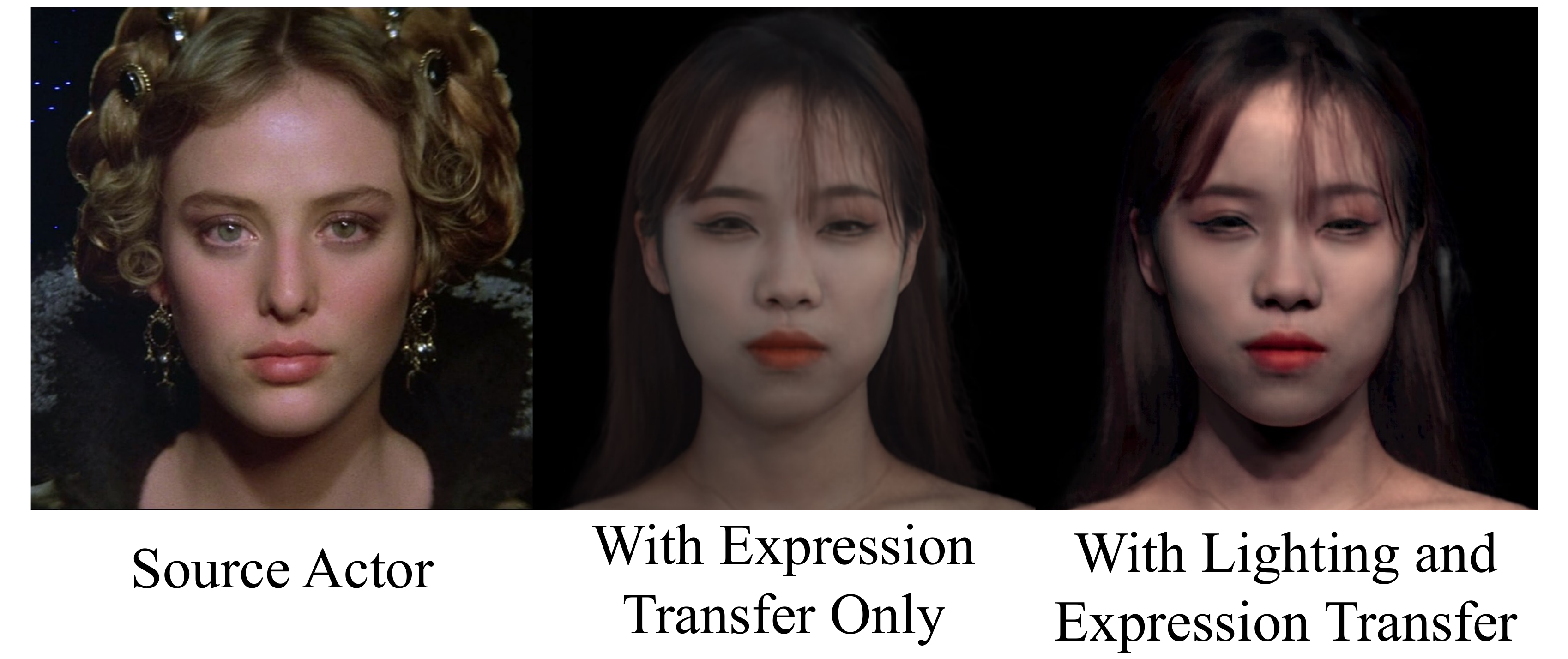}
  \caption{
    Application of virtual production.
  }
  \label{fig:film}
\end{figure}

The explicit disentanglement of the head pose, facial expression and lighting condition in NeRVP that produces a photo-realistic video portrait of the target actor enables many visual effects. 
Here, we introduce following basic operations of our approach, whose combination empowers various applications as demonstrated in Section~\ref{sec:results}.  

\textbf{Relighting.}
Since our approach synthesizes 4D reflectance fields for the target actor, we can relight the portrait video by linearly combining the RGB channels of OLAT images according to arbitrary target environment map. 
Such kind of explicit approach provides more reliable and realistic outputs than learning-based methods which easily produce low-contrast results.

\textbf{Matting and Background Replacement.}
We generated 4d reflectance fields to simulate the lighting on the side and back very well, which allows us to rim-lit the portrait by relighting method to make the contour of the portrait more distinguishable from the background. We use ModNet~\cite{MODNet} for matting of portraits, and we show the important role of rim-light in the matting process in Fig.~\ref{fig:matte}.

\textbf{Face reenactment.}
Given a video clip of the source actor, our approach can transfer the pose and the facial expression to the portrait video of target actor. 
%
%
Finally, the target actor's face in the synthesized portrait video will reenact the source actor. 


\section{Experimental Results}\label{sec:results}
In this section, we evaluate our approach in a variety of challenging scenarios.

\subsection{Qualitative result} 

\begin{figure}[t]
	\includegraphics[width=\linewidth]{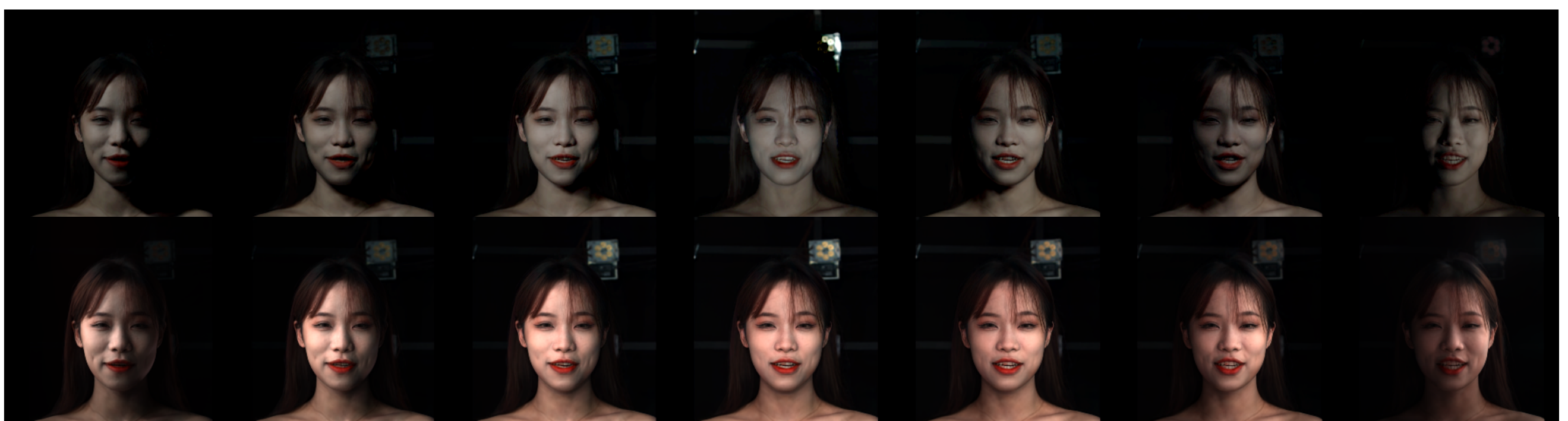}
	\caption{High frame rate dynamic OLAT generation.}
	\label{fig:OLAT}
\end{figure}


As we can see in the gallery Fig.~\ref{fig:gallery}, our approach can transfer the facial expression vividly. In addition, the network learns the OLAT frames explicitly, which makes it possible to relight the portraits in high-dynamic-range. This also enables us to imitate the photography studio lighting and show an artistic relighting effect. We chose Rembrandt lighting and cyberpunk lighting for demonstration. To imitate these two studio lighting, we need to have a good performance under various challenging lighting conditions, including rim light, hard directional light, backlight, top light and ambient light. Therefore, it is very difficult to imitate these two types of artistic lighting conditions, where our method shows superiority.


Our method also has many other applications. First, we can generate high frame rate dynamic OLAT. Since the network learns a mapping from facial parameters to reflection fields, we can interpolate lighting by interpolating expression parameters. This makes it possible to generate 1000fps OLAT data, which can be used in various scenes such as slow motion fill light in movies. In reality, a 100,000fps high-speed camera is required to collect such data, which cannot be achieved by current technology. We show this application in Fig.~\ref{fig:OLAT}. 

Our method can also be directly used in the field of virtual production, as illustrated in Fig.\ref{fig:film}. Our algorithm supports transferring such informative lighting and expressions from the given source actor to the target one, enabling film making with more flexibility out of the limitation of actors and surrounding illumination conditions.

\subsection{Comparisons}
\begin{figure}[t]
    \centering
	\includegraphics[width=0.8\linewidth]{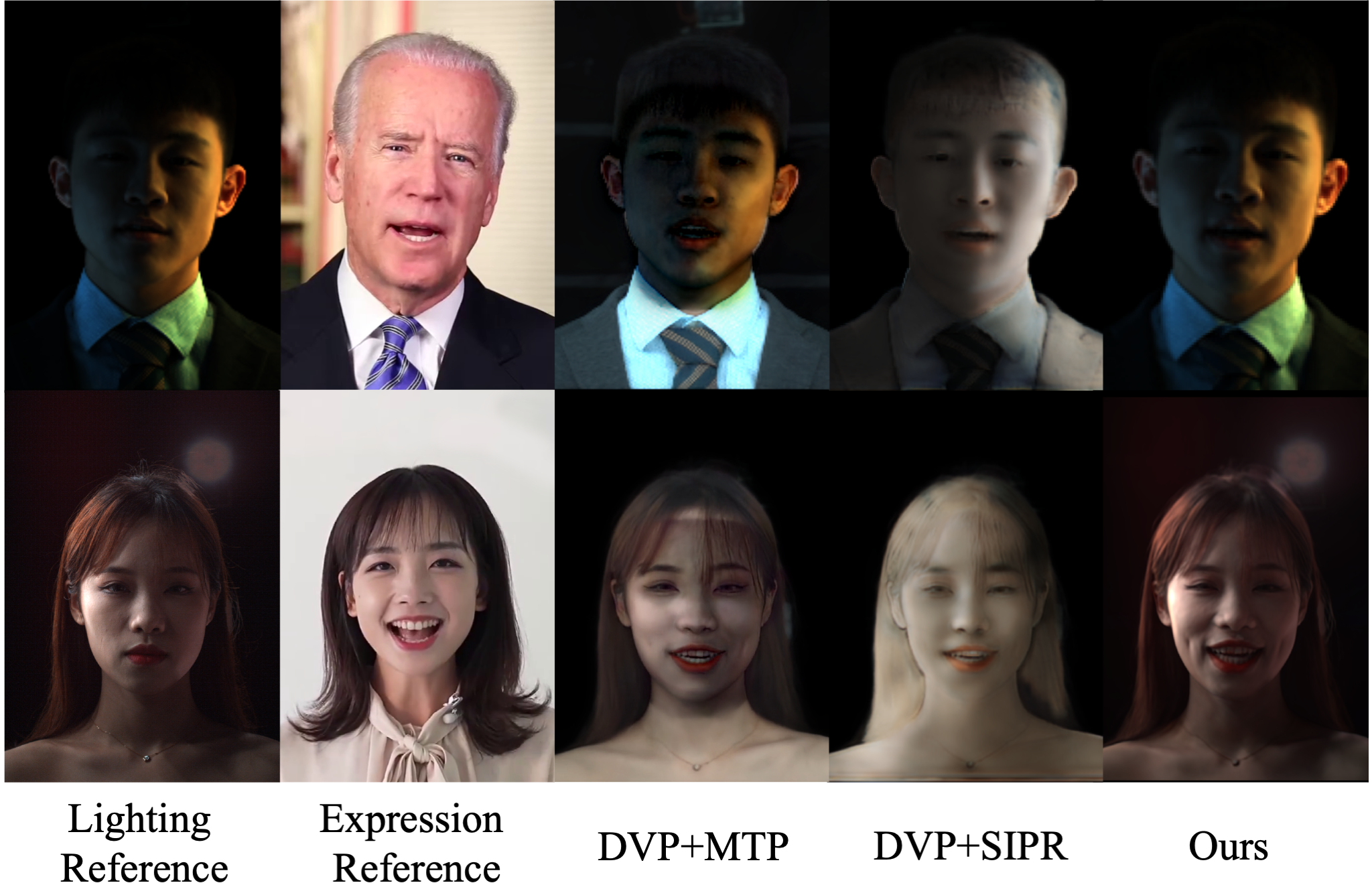}
	\caption{Qualitative comparison on in-the-wild videos.}
	\label{fig:comparison_test}
\end{figure}
With the aid of the immediate and high-quality dynamic facial reflectance field inference, our novel synthesized neural video portraits endow the entire capacity of combining dynamic facial expressions, various illumination conditions and changeable environmental backgrounds. It is particularly of competitive advantages over cases that relighting portraits are interfered by the interaction of the environment.
Since our method is a novel approach to generating relightable neural video portrait, to demonstrate our outstanding performance, we compare our method against the one based on DVP (Deep Video Portraits) ~\cite{kim2018deep}, which suffers a lack of relightable capacity. Therefore, for fair comparisons, we further add another state-of-the-art methods for relighting, Single Image Portrait Relight (SIPR) and MassTransport approach (MTP), respectively.

\begin{figure}[t]
    \centering
	\includegraphics[width=0.8\linewidth]{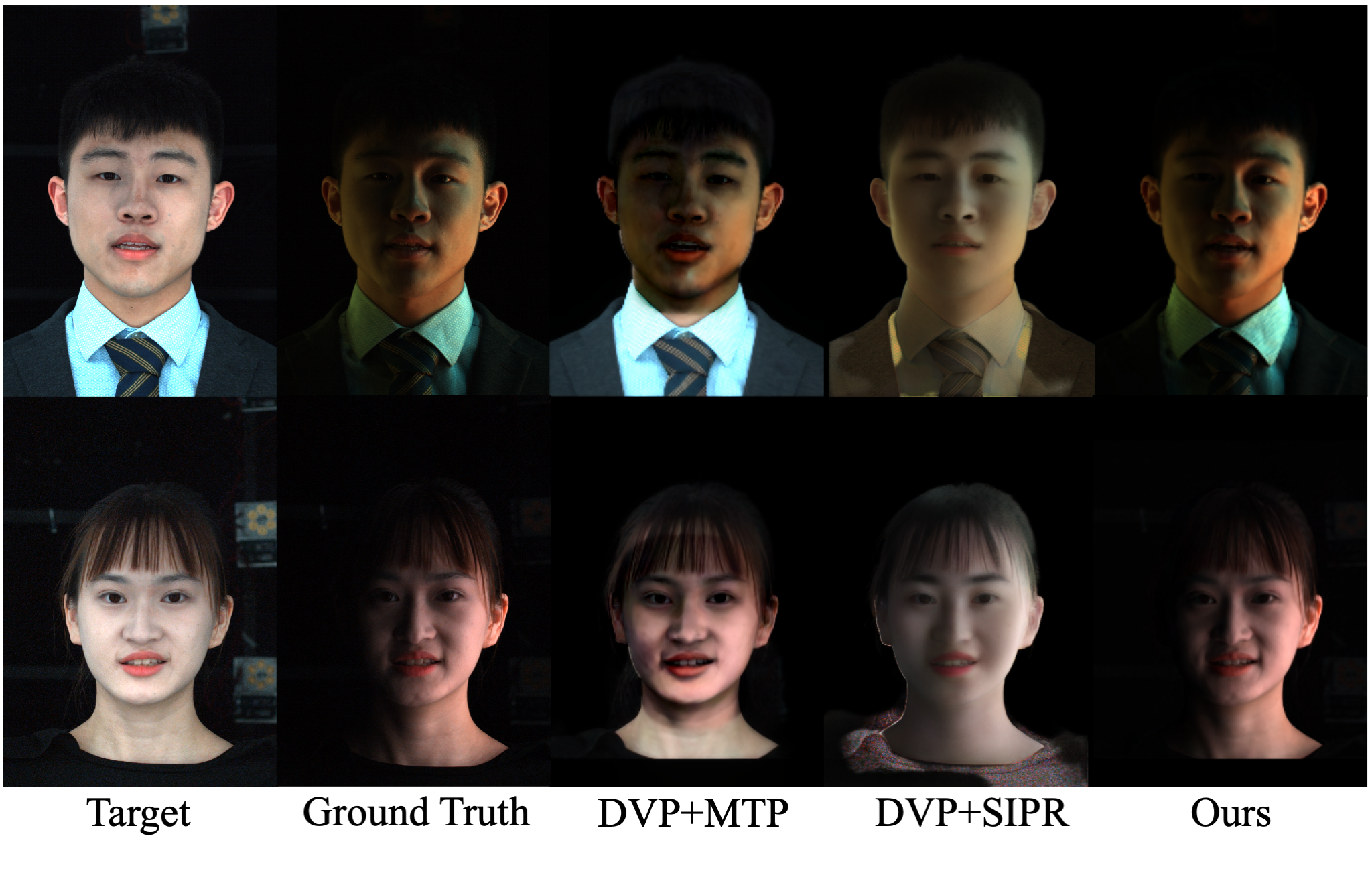}
	\caption{Qualitative comparison on OLAT test set.}
	\label{fig:comparison_train}
\end{figure}

As shown in both Fig.~\ref{fig:comparison_test} and Fig.~\ref{fig:comparison_train}, our approach surpasses all other methods in visual effects. DVP+SIR suffers from over-smoothness, leading to insufficiency of details and unevenly distributed light on the target portrait face. Moreover, DVP+MTP shows severe flickering along with varying illumination conditions from various angles of deflection.
In contrast, our approach achieves the perfect rendering result in terms of photo-realism without any artifact.

\begin{table}[tp]

\begin{tabular}{@{}lllll@{}}

\toprule
\multicolumn{5}{l}{Comparison with other relighting methods}                                         \\ \midrule
\multicolumn{2}{l}{Method}                              & PSNR($\uparrow$)        & SSIM ($\uparrow$)       & MAE ($\downarrow$)         \\
\multicolumn{2}{l}{DVP+MTP}                          & 19.299±1.709      & 0.702±0.056      & 0.064±0.009      \\
\multicolumn{2}{l}{DVP+SIPR}                    & 17.655±1.854      & 0.675±0.043      & 0.088±0.020      \\
\multicolumn{2}{l}{Ours}                    & \textbf{27.271±1.659}       & \textbf{0.862±0.016}      & \textbf{0.029±0.004}      \\ \bottomrule
\end{tabular}
\caption{Quantitative comparision. $\uparrow$ means the larger is better, while $\downarrow$ means smaller is better. In terms of three metrics, our method outperforms the others.}
\label{table:ablation}
\end{table}

For further quantitative comparison, we evaluate our rendering results via three various metrics: signal-to-noise ratio (\textbf{PSNR}), structural similarity index(\textbf{SSIM}) and mean absolute error (\textbf{MAE}) to compare with existing state-of-art approaches. 
We generate a sequence of relighted portraits from all reference views which are evenly spaced in range of illumination angles.
As shown in the Tab.~\ref{table:ablation}, our method shows the outstanding performance on the photo-realistic synthesis and controllable video editing in terms of all three metrics with even less standard deviation.
%
\begin{figure}[t]
    \centering
	\includegraphics[width=0.8\linewidth]{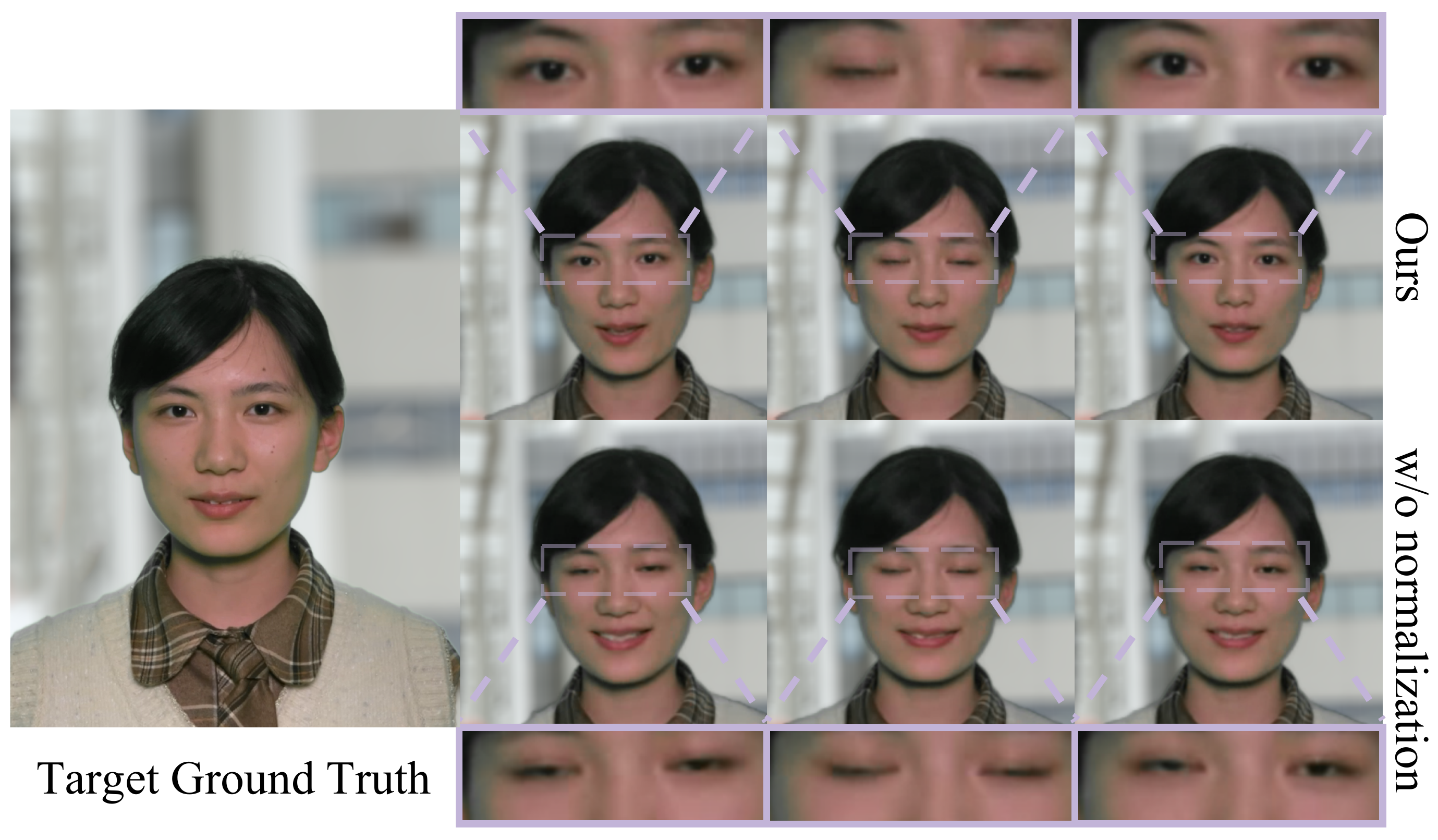}
	\caption{For the target person on the \textbf{left}, we use another person to animate. The results for our full pipeline are shown in the \textbf{first row}, and the \textbf{w/o normalization} test result is shown in the \textbf{second row}.}
	\label{fig:norm}
\end{figure}

\subsection{Ablation Study}
In this subsection, we evaluate the performance of our approach by comparing different designs of our method components in quality and quantity.
We use \textbf{w/o normalization} to denote the variations of our approach without normalizing the parameters of the extracted 3D face model.
As shown in Fig.~\ref{fig:norm}, the bottom row illustrates the blinking process without normalization on the parameters of the face model. In contrast, the upper row shows the reenactment of the same eye blinking motion after normalization. We can observe that when the source person's eyes are much smaller than the target person's, the animation results look like that the target person cannot open the eyes, which is absurd and unnatural. The key problem is that the source person and target person have different facial geometries, when we simply transfer the facial motions of the source person, the size of eyes and mouth will not match the face of the target person. Normalization maps the distribution of the parameters of the source person to the target person, which ensures that we can transfer facial motions in a reasonable and photo-realistic way.
%

\begin{figure}[t]
    \centering
	\includegraphics[width=0.8\linewidth]{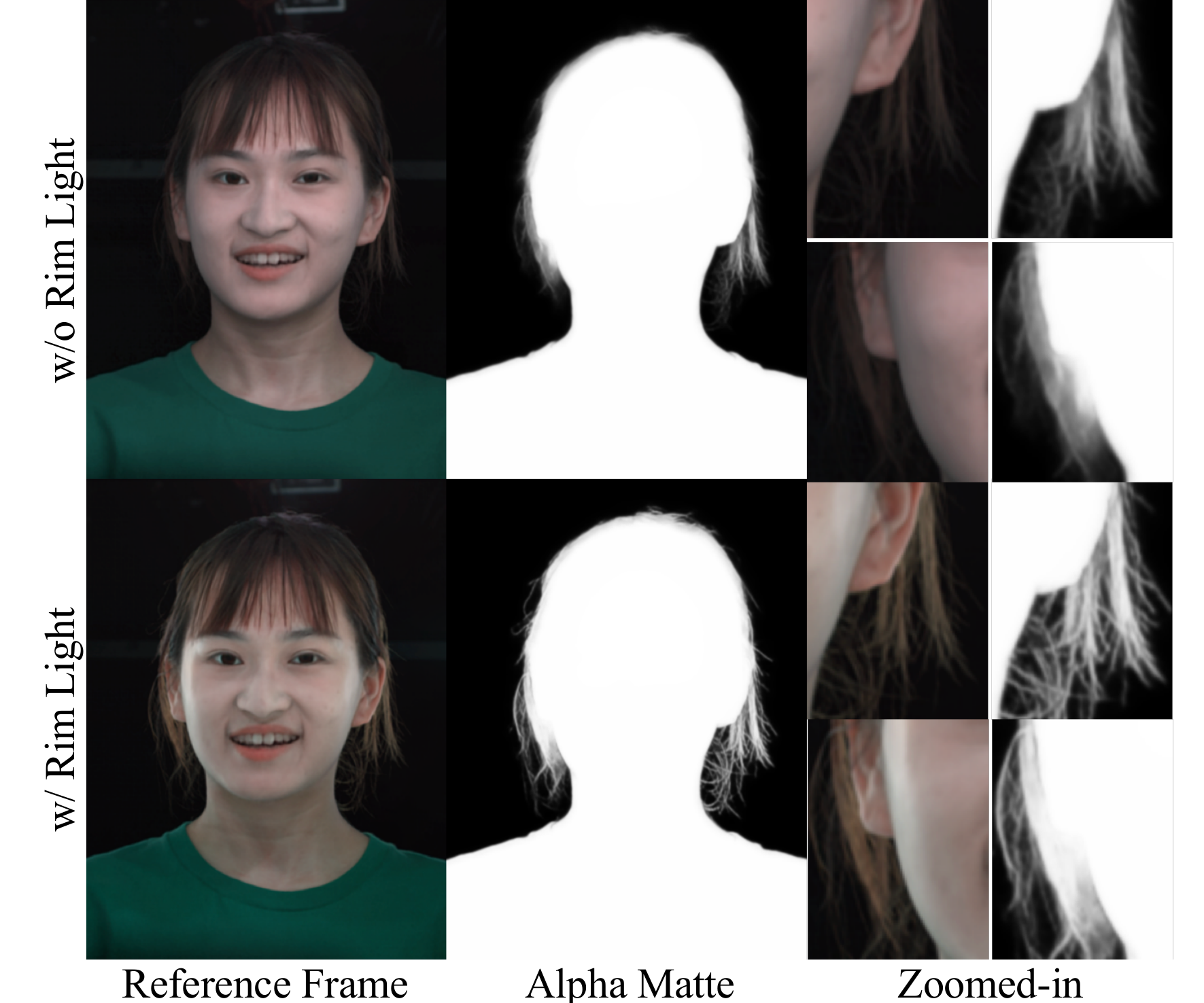}
	\caption{We showed the simulation of rim-light by generated neural reflectance field and demonstrated its importance in matting.}
	\label{fig:matte}
\end{figure}

When replacing the background, a good alpha map is necessary. We demonstrate the process of getting the alpha map in Fig.~\ref{fig:matte}
, in which we used the method of MODNet~\cite{MODNet}. We use \textbf{w/ rim-light} to denote using rim-light to lit the contour of the target actor, and use \textbf{w/o rim-light} to denote not using rim-light. The result shows that our generated neural reflectance field can simulate photorealistic rim-light, which greatly improves the quality of matting, especially for the fine structure such as hair.

\subsection{Limitations}
We have demonstrated the compelling capability of simultaneous relighting and reenactment for neural video portrait with a variety of fancy visual editing effects.
Nevertheless, as the first trial for such disentanglement of facial expression, background and lighting conditions for neural portrait generation, our approach is subject to some limitations.
%
%
%
First, the output of our neural network is 96 OLAT imagesets, leading to a huge GPU memory occupancy to support the training process. 
This is the key limitation on the spatial resolution of our generated results. 
Furthermore, to fully reenact the expression of the source person, we need to obtain a rich OLAT dataset of the target person, which is very expensive to train the person-specific neural network with a such a large-scale dataset.
%
%
%
%
In addition, our method is limited by the accuracy of the estimation of the facial parametric model, which means we cannot accurately estimate the representative exaggerated expression of each person. This makes our driving effect not accurate enough under some expressions.

%
Our approach also do not utilizes fine-grained 3D models for other regions such as hair. 
Thus, we can occasionally observe that the movements of these parts look unnatural and may violate the laws of physics.

\section{Conclusion}
We have presented the first approach to generate relightable neural video portraits of a target performer, which enables simultaneous relighting and reenactment via high-quality facial reflectance field learning.
Our novel pipeline, using multi-frame multi-task learning strategy,combines 4D reflectance field learning,  model-based facial performance capture and target-aware neural rendering, so as to transfer the  facial expressions from a source actor to a portrait video of a target actor with arbitrary new background
s and lighting conditions. 
%
%
%
Extensive experimental results demonstrate the effectiveness of our approach for simultaneous relighting and reenactment with various visual editing effects under high realism, which compares favorably to the state-of-the-art.
%
%
We believe that our approach is a critical step for the highly realistic synthesis of portrait video into various environments under the control of various meaningful attributes, with many potential applications for fancy visual effects in VR/AR, gaming, filming or entertainment.

{\small
\bibliographystyle{ieee_fullname}
\bibliography{ref}

\begin{thebibliography}{10}\itemsep=-1pt

\bibitem{Abrevaya_2020_CVPR}
Victoria~Fernandez Abrevaya, Adnane Boukhayma, Philip~H.S. Torr, and Edmond
  Boyer.
\newblock Cross-modal deep face normals with deactivable skip connections.
\newblock In {\em Proceedings of the IEEE/CVF Conference on Computer Vision and
  Pattern Recognition (CVPR)}, June 2020.

\bibitem{aldrian2012inverse}
Oswald Aldrian and William~AP Smith.
\newblock Inverse rendering of faces with a 3d morphable model.
\newblock {\em IEEE transactions on pattern analysis and machine intelligence},
  35(5):1080--1093, 2012.

\bibitem{averbuch2017bringing}
Hadar Averbuch-Elor, Daniel Cohen-Or, Johannes Kopf, and Michael~F Cohen.
\newblock Bringing portraits to life.
\newblock {\em ACM Transactions on Graphics (TOG)}, 36(6):1--13, 2017.

\bibitem{bi2021avatar}
Sai Bi, Stephen Lombardi, Shunsuke Saito, Tomas Simon, Shih-EnWei, Kevyn
  McPhail, Ravi Ramamoorthi, Yaser Sheikh, and Jason Saragih.
\newblock Deep relightable appearance models for animatable faces.
\newblock {\em ACM Trans. Graph. (Proc. SIGGRAPH)}, 40(4), 2021.

\bibitem{blanz2004exchanging}
Volker Blanz, Kristina Scherbaum, Thomas Vetter, and Hans-Peter Seidel.
\newblock Exchanging faces in images.
\newblock In {\em Computer Graphics Forum}, volume~23, pages 669--676. Wiley
  Online Library, 2004.

\bibitem{blanz1999morphable}
Volker Blanz and Thomas Vetter.
\newblock A morphable model for the synthesis of 3d faces.
\newblock In {\em Proceedings of the 26th annual conference on Computer
  graphics and interactive techniques}, pages 187--194, 1999.

\bibitem{cao2015real}
Chen Cao, Derek Bradley, Kun Zhou, and Thabo Beeler.
\newblock Real-time high-fidelity facial performance capture.
\newblock {\em ACM Transactions on Graphics (ToG)}, 34(4):1--9, 2015.

\bibitem{cao2014displaced}
Chen Cao, Qiming Hou, and Kun Zhou.
\newblock Displaced dynamic expression regression for real-time facial tracking
  and animation.
\newblock {\em ACM Transactions on graphics (TOG)}, 33(4):1--10, 2014.

\bibitem{chen2020free}
Anpei Chen, Ruiyang Liu, Ling Xie, and Jingyi Yu.
\newblock A free viewpoint portrait generator with dynamic styling.
\newblock {\em arXiv preprint arXiv:2007.03780}, 2020.

\bibitem{chen2011face}
Xiaowu Chen, Mengmeng Chen, Xin Jin, and Qinping Zhao.
\newblock Face illumination transfer through edge-preserving filters.
\newblock In {\em CVPR 2011}, pages 281--287. IEEE, 2011.

\bibitem{debevec2000acquiring}
Paul Debevec, Tim Hawkins, Chris Tchou, Haarm-Pieter Duiker, Westley Sarokin,
  and Mark Sagar.
\newblock Acquiring the reflectance field of a human face.
\newblock In {\em Proceedings of the 27th annual conference on Computer
  graphics and interactive techniques}, pages 145--156, 2000.

\bibitem{DBLP2018patch}
Ugur Demir and G{\"{o}}zde~B. {\"{U}}nal.
\newblock Patch-based image inpainting with generative adversarial networks.
\newblock {\em CoRR}, abs/1803.07422, 2018.

\bibitem{egger2018occlusion}
Bernhard Egger, Sandro Sch{\"o}nborn, Andreas Schneider, Adam Kortylewski,
  Andreas Morel-Forster, Clemens Blumer, and Thomas Vetter.
\newblock Occlusion-aware 3d morphable models and an illumination prior for
  face image analysis.
\newblock {\em International Journal of Computer Vision}, 126(12):1269--1287,
  2018.

\bibitem{egger20203d}
Bernhard Egger, William~AP Smith, Ayush Tewari, Stefanie Wuhrer, Michael
  Zollhoefer, Thabo Beeler, Florian Bernard, Timo Bolkart, Adam Kortylewski,
  Sami Romdhani, et~al.
\newblock 3d morphable face models—past, present, and future.
\newblock {\em ACM Transactions on Graphics (TOG)}, 39(5):1--38, 2020.

\bibitem{feng2020learning}
Yao Feng, Haiwen Feng, Michael~J Black, and Timo Bolkart.
\newblock Learning an animatable detailed 3d face model from in-the-wild
  images.
\newblock {\em arXiv preprint arXiv:2012.04012}, 2020.

\bibitem{fried2019text}
Ohad Fried, Ayush Tewari, Michael Zollh{\"o}fer, Adam Finkelstein, Eli
  Shechtman, Dan~B Goldman, Kyle Genova, Zeyu Jin, Christian Theobalt, and
  Maneesh Agrawala.
\newblock Text-based editing of talking-head video.
\newblock {\em ACM Transactions on Graphics (TOG)}, 38(4):1--14, 2019.

\bibitem{fyffe2014driving}
Graham Fyffe, Andrew Jones, Oleg Alexander, Ryosuke Ichikari, and Paul Debevec.
\newblock Driving high-resolution facial scans with video performance capture.
\newblock {\em ACM Transactions on Graphics (TOG)}, 34(1):1--14, 2014.

\bibitem{gafni2020dynamic}
Guy Gafni, Justus Thies, Michael Zollh{\"o}fer, and Matthias Nie{\ss}ner.
\newblock Dynamic neural radiance fields for monocular 4d facial avatar
  reconstruction.
\newblock {\em arXiv preprint arXiv:2012.03065}, 2020.

\bibitem{garrido2014automatic}
Pablo Garrido, Levi Valgaerts, Ole Rehmsen, Thorsten Thormahlen, Patrick Perez,
  and Christian Theobalt.
\newblock Automatic face reenactment.
\newblock In {\em Proceedings of the IEEE conference on computer vision and
  pattern recognition}, pages 4217--4224, 2014.

\bibitem{garrido2016reconstruction}
Pablo Garrido, Michael Zollh{\"o}fer, Dan Casas, Levi Valgaerts, Kiran
  Varanasi, Patrick P{\'e}rez, and Christian Theobalt.
\newblock Reconstruction of personalized 3d face rigs from monocular video.
\newblock {\em ACM Transactions on Graphics (TOG)}, 35(3):1--15, 2016.

\bibitem{guo2019relightables}
Kaiwen Guo, Peter Lincoln, Philip Davidson, Jay Busch, Xueming Yu, Matt Whalen,
  Geoff Harvey, Sergio Orts-Escolano, Rohit Pandey, Jason Dourgarian, et~al.
\newblock The relightables: Volumetric performance capture of humans with
  realistic relighting.
\newblock {\em ACM Transactions on Graphics (TOG)}, 38(6):1--19, 2019.

\bibitem{ichim2015dynamic}
Alexandru~Eugen Ichim, Sofien Bouaziz, and Mark Pauly.
\newblock Dynamic 3d avatar creation from hand-held video input.
\newblock {\em ACM Transactions on Graphics (ToG)}, 34(4):1--14, 2015.

\bibitem{isola2017image}
Phillip Isola, Jun-Yan Zhu, Tinghui Zhou, and Alexei~A Efros.
\newblock Image-to-image translation with conditional adversarial networks.
\newblock In {\em Proceedings of the IEEE conference on computer vision and
  pattern recognition}, pages 1125--1134, 2017.

\bibitem{ji2021audio}
Xinya Ji, Hang Zhou, Kaisiyuan Wang, Wayne Wu, Chen~Change Loy, Xun Cao, and
  Feng Xu.
\newblock Audio-driven emotional video portraits.
\newblock {\em arXiv preprint arXiv:2104.07452}, 2021.

\bibitem{karras2019style}
Tero Karras, Samuli Laine, and Timo Aila.
\newblock A style-based generator architecture for generative adversarial
  networks.
\newblock In {\em Proceedings of the IEEE/CVF Conference on Computer Vision and
  Pattern Recognition}, pages 4401--4410, 2019.

\bibitem{kazemi2014one}
Vahid Kazemi and Josephine Sullivan.
\newblock One millisecond face alignment with an ensemble of regression trees.
\newblock In {\em Proceedings of the IEEE conference on computer vision and
  pattern recognition}, pages 1867--1874, 2014.

\bibitem{MODNet}
Zhanghan Ke, Kaican Li, Yurou Zhou, Qiuhua Wu, Xiangyu Mao, Qiong Yan, and
  Rynson~W.H. Lau.
\newblock Is a green screen really necessary for real-time portrait matting?
\newblock {\em ArXiv}, abs/2011.11961, 2020.

\bibitem{kemelmacher2013internet}
Ira Kemelmacher-Shlizerman.
\newblock Internet based morphable model.
\newblock In {\em Proceedings of the IEEE international conference on computer
  vision}, pages 3256--3263, 2013.

\bibitem{kemelmacher2010being}
Ira Kemelmacher-Shlizerman, Aditya Sankar, Eli Shechtman, and Steven~M Seitz.
\newblock Being john malkovich.
\newblock In {\em European Conference on Computer Vision}, pages 341--353.
  Springer, 2010.

\bibitem{kim2019neural}
Hyeongwoo Kim, Mohamed Elgharib, Michael Zollh{\"o}fer, Hans-Peter Seidel,
  Thabo Beeler, Christian Richardt, and Christian Theobalt.
\newblock Neural style-preserving visual dubbing.
\newblock {\em ACM Transactions on Graphics (TOG)}, 38(6):1--13, 2019.

\bibitem{kim2018deep}
Hyeongwoo Kim, Pablo Garrido, Ayush Tewari, Weipeng Xu, Justus Thies, Matthias
  Niessner, Patrick P{\'e}rez, Christian Richardt, Michael Zollh{\"o}fer, and
  Christian Theobalt.
\newblock Deep video portraits.
\newblock {\em ACM Transactions on Graphics (TOG)}, 37(4):1--14, 2018.

\bibitem{li2013data}
Kai Li, Qionghai Dai, Ruiping Wang, Yebin Liu, Feng Xu, and Jue Wang.
\newblock A data-driven approach for facial expression retargeting in video.
\newblock {\em IEEE Transactions on Multimedia}, 16(2):299--310, 2013.

\bibitem{li2017learning}
Tianye Li, Timo Bolkart, Michael~J Black, Hao Li, and Javier Romero.
\newblock Learning a model of facial shape and expression from 4d scans.
\newblock {\em ACM Trans. Graph.}, 36(6):194--1, 2017.

\bibitem{liu2011realistic}
Kang Liu and Joern Ostermann.
\newblock Realistic facial expression synthesis for an image-based talking
  head.
\newblock In {\em 2011 IEEE International Conference on Multimedia and Expo},
  pages 1--6. IEEE, 2011.

\bibitem{liu2021relighting}
Yang Liu, Alexandros Neophytou, Sunando Sengupta, and Eric Sommerlade.
\newblock Relighting images in the wild with a self-supervised siamese
  auto-encoder.
\newblock In {\em Proceedings of the IEEE/CVF Winter Conference on Applications
  of Computer Vision}, pages 32--40, 2021.

\bibitem{lombardi2018deep}
Stephen Lombardi, Jason Saragih, Tomas Simon, and Yaser Sheikh.
\newblock Deep appearance models for face rendering.
\newblock {\em ACM Transactions on Graphics (TOG)}, 37(4):1--13, 2018.

\bibitem{meka2019deep}
Abhimitra Meka, Christian Haene, Rohit Pandey, Michael Zollh{\"o}fer, Sean
  Fanello, Graham Fyffe, Adarsh Kowdle, Xueming Yu, Jay Busch, Jason
  Dourgarian, et~al.
\newblock Deep reflectance fields: high-quality facial reflectance field
  inference from color gradient illumination.
\newblock {\em ACM Transactions on Graphics (TOG)}, 38(4):1--12, 2019.

\bibitem{meka2020deep}
Abhimitra Meka, Rohit Pandey, Christian H{\"a}ne, Sergio Orts-Escolano, Peter
  Barnum, Philip David-Son, Daniel Erickson, Yinda Zhang, Jonathan Taylor,
  Sofien Bouaziz, et~al.
\newblock Deep relightable textures: volumetric performance capture with neural
  rendering.
\newblock {\em ACM Transactions on Graphics (TOG)}, 39(6):1--21, 2020.

\bibitem{nestmeyer2020learning}
Thomas Nestmeyer, Jean-Fran{\c{c}}ois Lalonde, Iain Matthews, and Andreas
  Lehrmann.
\newblock Learning physics-guided face relighting under directional light.
\newblock In {\em Proceedings of the IEEE/CVF Conference on Computer Vision and
  Pattern Recognition}, pages 5124--5133, 2020.

\bibitem{pandey2021total}
Rohit~Kumar Pandey, Sergio~Orts Escolano, Chloe LeGendre, Christian Haene,
  Sofien Bouaziz, Christoph Rhemann, Paul Debevec, and Sean Fanello.
\newblock Total relighting: Learning to relight portraits for background
  replacement.
\newblock 2021.

\bibitem{richardson20163d}
Elad Richardson, Matan Sela, and Ron Kimmel.
\newblock 3d face reconstruction by learning from synthetic data.
\newblock In {\em 2016 fourth international conference on 3D vision (3DV)},
  pages 460--469. IEEE, 2016.

\bibitem{richardson2017learning}
Elad Richardson, Matan Sela, Roy Or-El, and Ron Kimmel.
\newblock Learning detailed face reconstruction from a single image.
\newblock In {\em Proceedings of the IEEE conference on computer vision and
  pattern recognition}, pages 1259--1268, 2017.

\bibitem{roth2016adaptive}
Joseph Roth, Yiying Tong, and Xiaoming Liu.
\newblock Adaptive 3d face reconstruction from unconstrained photo collections.
\newblock In {\em Proceedings of the IEEE Conference on Computer Vision and
  Pattern Recognition}, pages 4197--4206, 2016.

\bibitem{sengupta2018sfsnet}
Soumyadip Sengupta, Angjoo Kanazawa, Carlos~D Castillo, and David~W Jacobs.
\newblock Sfsnet: Learning shape, reflectance and illuminance of facesin the
  wild'.
\newblock In {\em Proceedings of the IEEE conference on computer vision and
  pattern recognition}, pages 6296--6305, 2018.

\bibitem{shih2014style}
YiChang Shih, Sylvain Paris, Connelly Barnes, William~T Freeman, and Fr{\'e}do
  Durand.
\newblock Style transfer for headshot portraits.
\newblock 2014.

\bibitem{shu2017portrait}
Zhixin Shu, Sunil Hadap, Eli Shechtman, Kalyan Sunkavalli, Sylvain Paris, and
  Dimitris Samaras.
\newblock Portrait lighting transfer using a mass transport approach.
\newblock {\em ACM Transactions on Graphics (TOG)}, 36(4):1, 2017.

\bibitem{song2017stylizing}
Yibing Song, Linchao Bao, Shengfeng He, Qingxiong Yang, and Ming-Hsuan Yang.
\newblock Stylizing face images via multiple exemplars.
\newblock {\em Computer Vision and Image Understanding}, 162:135--145, 2017.

\bibitem{sun2019single}
Tiancheng Sun, Jonathan~T Barron, Yun-Ta Tsai, Zexiang Xu, Xueming Yu, Graham
  Fyffe, Christoph Rhemann, Jay Busch, Paul~E Debevec, and Ravi Ramamoorthi.
\newblock Single image portrait relighting.
\newblock {\em ACM Trans. Graph.}, 38(4):79--1, 2019.

\bibitem{sun2020light}
Tiancheng Sun, Zexiang Xu, Xiuming Zhang, Sean Fanello, Christoph Rhemann, Paul
  Debevec, Yun-Ta Tsai, Jonathan~T Barron, and Ravi Ramamoorthi.
\newblock Light stage super-resolution: continuous high-frequency relighting.
\newblock {\em ACM Transactions on Graphics (TOG)}, 39(6):1--12, 2020.

\bibitem{suwajanakorn2014total}
Supasorn Suwajanakorn, Ira Kemelmacher-Shlizerman, and Steven~M Seitz.
\newblock Total moving face reconstruction.
\newblock In {\em European conference on computer vision}, pages 796--812.
  Springer, 2014.

\bibitem{suwajanakorn2015makes}
Supasorn Suwajanakorn, Steven~M Seitz, and Ira Kemelmacher-Shlizerman.
\newblock What makes tom hanks look like tom hanks.
\newblock In {\em Proceedings of the IEEE International Conference on Computer
  Vision}, pages 3952--3960, 2015.

\bibitem{tewari2019fml}
Ayush Tewari, Florian Bernard, Pablo Garrido, Gaurav Bharaj, Mohamed Elgharib,
  Hans-Peter Seidel, Patrick P{\'e}rez, Michael Zollhofer, and Christian
  Theobalt.
\newblock Fml: Face model learning from videos.
\newblock In {\em Proceedings of the IEEE/CVF Conference on Computer Vision and
  Pattern Recognition}, pages 10812--10822, 2019.

\bibitem{tewari2021photoapp}
Ayush Tewari, Abdallah Dib, Tim Weyrich, Bernd Bickel, Hans-Peter Seidel,
  Hanspeter Pfister, Wojciech Matusik, Louis Chevallier, Mohamed Elgharib,
  Christian Theobalt, et~al.
\newblock Photoapp: Photorealistic appearance editing of head portraits.
\newblock {\em arXiv preprint arXiv:2103.07658}, 2021.

\bibitem{tewari2020pie}
Ayush Tewari, Mohamed Elgharib, Florian Bernard, Hans-Peter Seidel, Patrick
  P{\'e}rez, Michael Zollh{\"o}fer, and Christian Theobalt.
\newblock Pie: Portrait image embedding for semantic control.
\newblock {\em ACM Transactions on Graphics (TOG)}, 39(6):1--14, 2020.

\bibitem{tewari2020stylerig}
Ayush Tewari, Mohamed Elgharib, Gaurav Bharaj, Florian Bernard, Hans-Peter
  Seidel, Patrick P{\'e}rez, Michael Zollhofer, and Christian Theobalt.
\newblock Stylerig: Rigging stylegan for 3d control over portrait images.
\newblock In {\em Proceedings of the IEEE/CVF Conference on Computer Vision and
  Pattern Recognition}, pages 6142--6151, 2020.

\bibitem{NR_survey}
Ayush Tewari, Ohad Fried, Justus Thies, Vincent Sitzmann, Stephen Lombardi,
  Kalyan Sunkavalli, Ricardo Martin-Brualla, Tomas Simon, Jason Saragih,
  Matthias Nießner, Rohit Pandey, Sean Fanello, Gordon Wetzstein, Jun-Yan Zhu,
  Christian Theobalt, Maneesh Agrawala, Eli Shechtman, Dan~B. Goldman, and
  Michael Zollhöfer.
\newblock {State of the Art on Neural Rendering}.
\newblock {\em Computer Graphics Forum}, 2020.

\bibitem{tewari2020monocular}
Ayush Tewari, Tae-Hyun Oh, Tim Weyrich, Bernd Bickel, Hans-Peter Seidel,
  Hanspeter Pfister, Wojciech Matusik, Mohamed Elgharib, Christian Theobalt,
  et~al.
\newblock Monocular reconstruction of neural face reflectance fields.
\newblock {\em arXiv preprint arXiv:2008.10247}, 2020.

\bibitem{tewari2018self}
Ayush Tewari, Michael Zollh{\"o}fer, Pablo Garrido, Florian Bernard, Hyeongwoo
  Kim, Patrick P{\'e}rez, and Christian Theobalt.
\newblock Self-supervised multi-level face model learning for monocular
  reconstruction at over 250 hz.
\newblock In {\em Proceedings of the IEEE Conference on Computer Vision and
  Pattern Recognition}, pages 2549--2559, 2018.

\bibitem{tewari2017mofa}
Ayush Tewari, Michael Zollhofer, Hyeongwoo Kim, Pablo Garrido, Florian Bernard,
  Patrick Perez, and Christian Theobalt.
\newblock Mofa: Model-based deep convolutional face autoencoder for
  unsupervised monocular reconstruction.
\newblock In {\em Proceedings of the IEEE International Conference on Computer
  Vision Workshops}, pages 1274--1283, 2017.

\bibitem{thies2020neural}
Justus Thies, Mohamed Elgharib, Ayush Tewari, Christian Theobalt, and Matthias
  Nie{\ss}ner.
\newblock Neural voice puppetry: Audio-driven facial reenactment.
\newblock In {\em European Conference on Computer Vision}, pages 716--731.
  Springer, 2020.

\bibitem{thies2019deferred}
Justus Thies, Michael Zollh{\"o}fer, and Matthias Nie{\ss}ner.
\newblock Deferred neural rendering: Image synthesis using neural textures.
\newblock {\em ACM Transactions on Graphics (TOG)}, 38(4):1--12, 2019.

\bibitem{thies2015real}
Justus Thies, Michael Zollh{\"o}fer, Matthias Nie{\ss}ner, Levi Valgaerts, Marc
  Stamminger, and Christian Theobalt.
\newblock Real-time expression transfer for facial reenactment.
\newblock {\em ACM Trans. Graph.}, 34(6):183--1, 2015.

\bibitem{thies2016face2face}
Justus Thies, Michael Zollhofer, Marc Stamminger, Christian Theobalt, and
  Matthias Nie{\ss}ner.
\newblock Face2face: Real-time face capture and reenactment of rgb videos.
\newblock In {\em Proceedings of the IEEE conference on computer vision and
  pattern recognition}, pages 2387--2395, 2016.

\bibitem{thies2018facevr}
Justus Thies, Michael Zollh{\"o}fer, Marc Stamminger, Christian Theobalt, and
  Matthias Nie{\ss}ner.
\newblock Facevr: Real-time gaze-aware facial reenactment in virtual reality.
\newblock {\em ACM Transactions on Graphics (TOG)}, 37(2):1--15, 2018.

\bibitem{thies2018headon}
Justus Thies, Michael Zollh{\"o}fer, Christian Theobalt, Marc Stamminger, and
  Matthias Nie{\ss}ner.
\newblock Headon: Real-time reenactment of human portrait videos.
\newblock {\em ACM Transactions on Graphics (TOG)}, 37(4):1--13, 2018.

\bibitem{vlasic2006face}
Daniel Vlasic, Matthew Brand, Hanspeter Pfister, and Jovan Popovic.
\newblock Face transfer with multilinear models.
\newblock In {\em ACM SIGGRAPH 2006 Courses}, pages 24--es. 2006.

\bibitem{wang2007face}
Yang Wang, Zicheng Liu, Gang Hua, Zhen Wen, Zhengyou Zhang, and Dimitris
  Samaras.
\newblock Face re-lighting from a single image under harsh lighting conditions.
\newblock In {\em 2007 IEEE Conference on Computer Vision and Pattern
  Recognition}, pages 1--8. IEEE, 2007.

\bibitem{1284395}
Zhou Wang, A.C. Bovik, H.R. Sheikh, and E.P. Simoncelli.
\newblock Image quality assessment: from error visibility to structural
  similarity.
\newblock {\em IEEE Transactions on Image Processing}, 13(4):600--612, 2004.

\bibitem{wang2020single}
Zhibo Wang, Xin Yu, Ming Lu, Quan Wang, Chen Qian, and Feng Xu.
\newblock Single image portrait relighting via explicit multiple reflectance
  channel modeling.
\newblock {\em ACM Transactions on Graphics (TOG)}, 39(6):1--13, 2020.

\bibitem{wright2006digital}
Steve Wright.
\newblock {\em Digital compositing for film and video}.
\newblock Routledge, 2006.

\bibitem{wu2016anatomically}
Chenglei Wu, Derek Bradley, Markus Gross, and Thabo Beeler.
\newblock An anatomically-constrained local deformation model for monocular
  face capture.
\newblock {\em ACM transactions on graphics (TOG)}, 35(4):1--12, 2016.

\bibitem{yamaguchi2018high}
Shugo Yamaguchi, Shunsuke Saito, Koki Nagano, Yajie Zhao, Weikai Chen, Kyle
  Olszewski, Shigeo Morishima, and Hao Li.
\newblock High-fidelity facial reflectance and geometry inference from an
  unconstrained image.
\newblock {\em ACM Transactions on Graphics (TOG)}, 37(4):1--14, 2018.

\bibitem{yao2020talkinghead}
Xinwei Yao, Ohad Fried, Kayvon Fatahalian, and Maneesh Agrawala.
\newblock Iterative text-based editing of talking-heads using neural
  retargeting, 2020.

\bibitem{zhang2021neural}
Xiuming Zhang, Sean Fanello, Yun-Ta Tsai, Tiancheng Sun, Tianfan Xue, Rohit
  Pandey, Sergio Orts-Escolano, Philip Davidson, Christoph Rhemann, Paul
  Debevec, et~al.
\newblock Neural light transport for relighting and view synthesis.
\newblock {\em ACM Transactions on Graphics (TOG)}, 40(1):1--17, 2021.

\bibitem{zhou2019deep}
Hao Zhou, Sunil Hadap, Kalyan Sunkavalli, and David~W Jacobs.
\newblock Deep single-image portrait relighting.
\newblock In {\em Proceedings of the IEEE/CVF International Conference on
  Computer Vision}, pages 7194--7202, 2019.

\bibitem{zhu2017face}
Xiangyu Zhu, Xiaoming Liu, Zhen Lei, and Stan~Z Li.
\newblock Face alignment in full pose range: A 3d total solution.
\newblock {\em IEEE transactions on pattern analysis and machine intelligence},
  41(1):78--92, 2017.

\bibitem{zollhofer2018state}
Michael Zollh{\"o}fer, Justus Thies, Pablo Garrido, Derek Bradley, Thabo
  Beeler, Patrick P{\'e}rez, Marc Stamminger, Matthias Nie{\ss}ner, and
  Christian Theobalt.
\newblock State of the art on monocular 3d face reconstruction, tracking, and
  applications.
\newblock In {\em Computer Graphics Forum}, volume~37, pages 523--550. Wiley
  Online Library, 2018.

\end{thebibliography}
}

\title{Supplementary Materials}

\maketitle

\begin{figure}[t]
 \includegraphics[width=\linewidth]{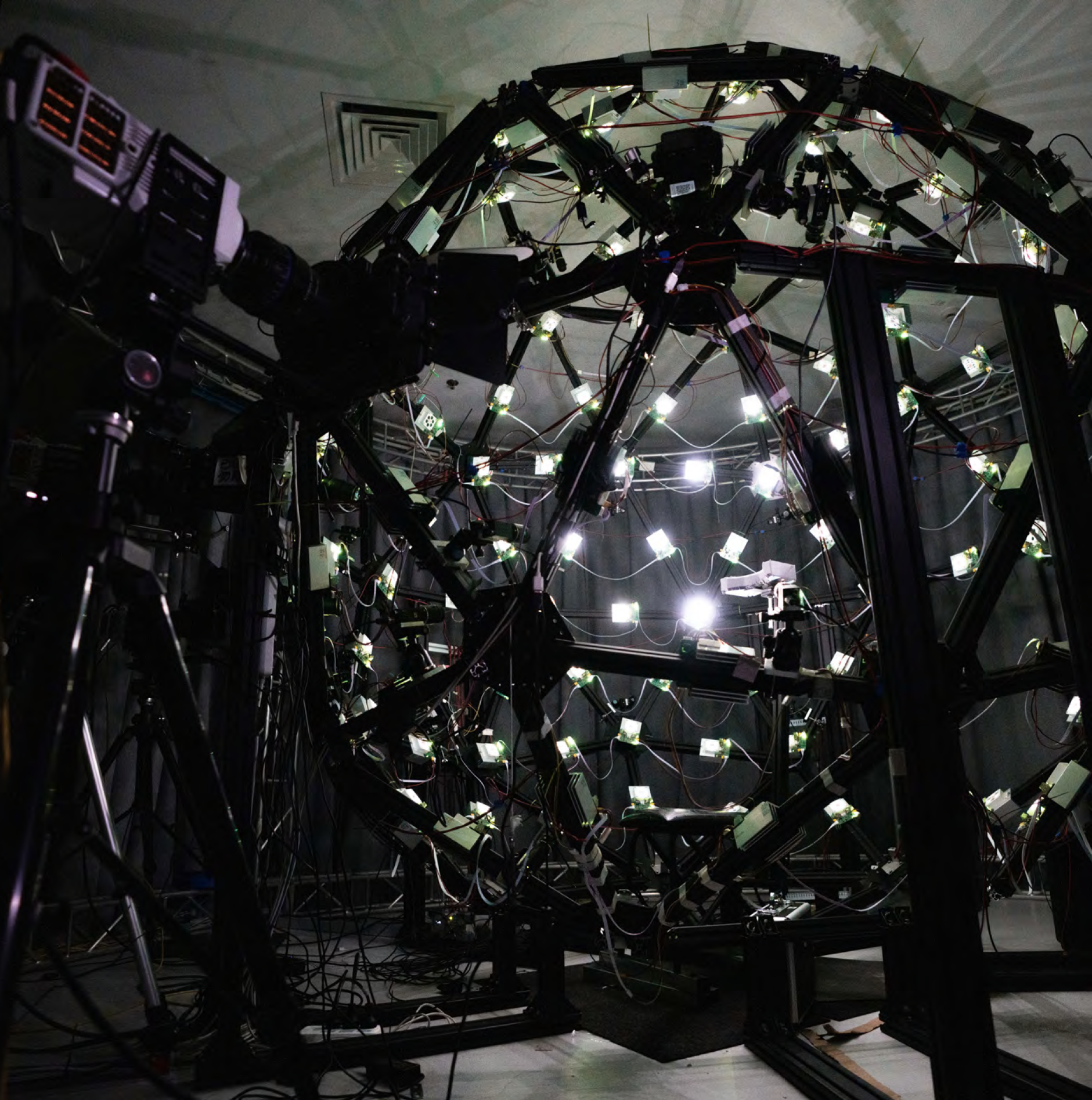}
 \caption{The demonstration of our capturing system. Cameras and lights are arranged on a spherical structure.}
 \label{fig:capture_system}
\end{figure}
\section{Hardware architecture and capture settings}
To recover the 4D reflectance fields of dynamic portrait, we build up a light stage device with 96 LED light sources and a stationary 4K ultra-high-speed camera at 1000 fps, resulting in a temporal OLAT image set at 25 fps, so as to provide fine-grained facial perception.

Under such a dynamic capture setting, the target performer can freely speak, translate and rotate in a certain range to provide a continuous and dynamic OLAT image sets sequence. 
However, one of the most challenging issues is that the motion of the captured target along with data acquisition will cause misalignment, leading to blurriness in the OLAT image sets, making the data post-processing more challenging. 
We conquer such limitations using an optical flow algorithm and further retrieve results at a higher frame rate using densely-acquired homogeneous full-lit frames.

Our hardware architecture is demonstrated in Fig.~\ref{fig:capture_system}, which is a spherical dome of a radius of 1.3 meters with 96 fully programmable LEDs and a 4K ultra-high-speed camera.
The LEDs are independently controlled via a synchronization control system and evenly localized on the doom to provide smooth lighting conditions. 

Single PCC (Phantom Capture Camera) is leveraged with a global shutter setting, generating roughly 6 TB of data in total. 
For each data session, we collect video sequences in 25 seconds rather than isolated frames with 700 us exposure time and 1000 EI, which allows a lower noise floor. 
In practice, we can simultaneously control the PCC along with the LEDs system. 

\begin{figure}[t]
\centering
 \includegraphics[width=0.85\linewidth]{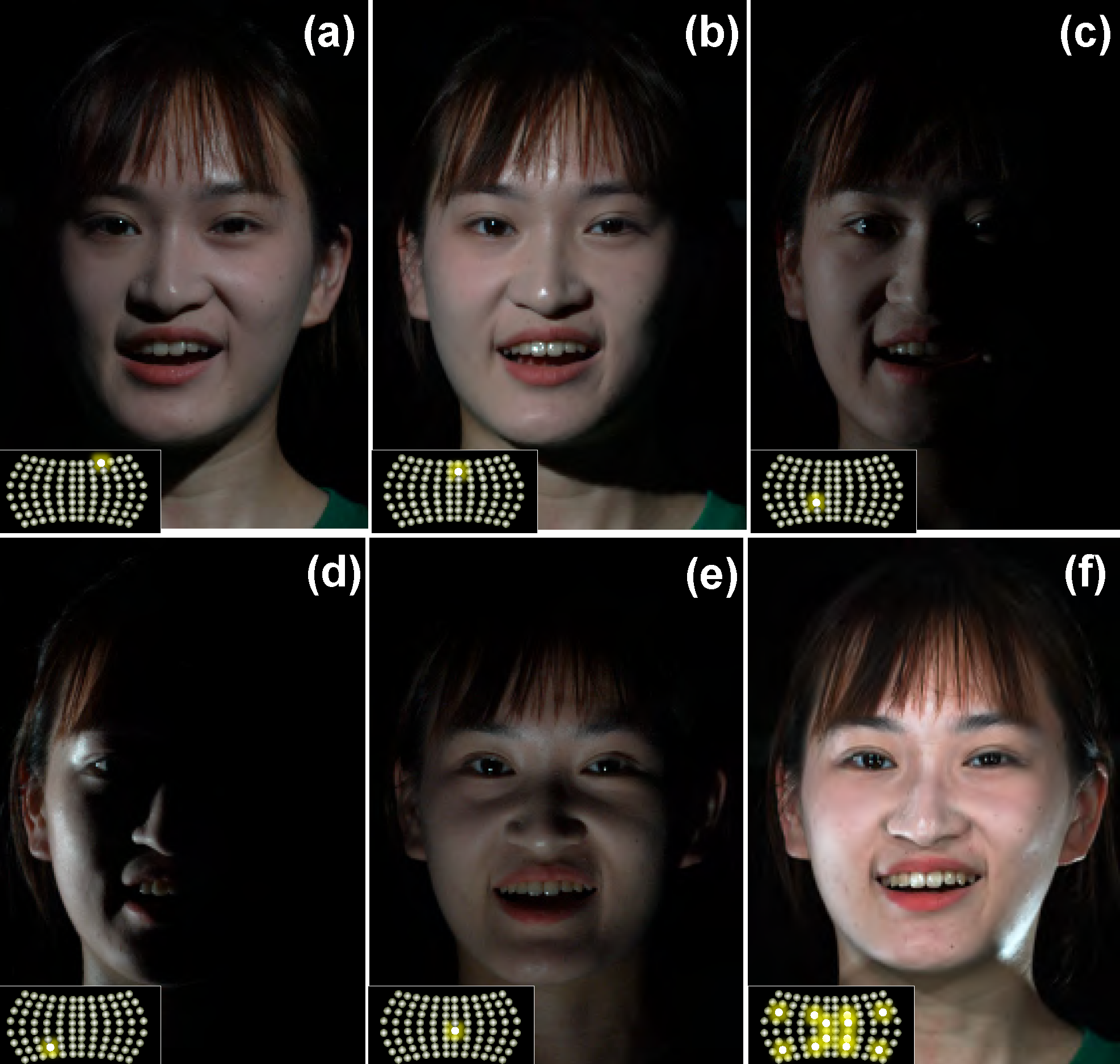}
 \caption{The samples of our captured data. (a) to (e) are samples of OLAT images; (f) is a full-lit frame image for calculating the optical-flow. The bottom-left illustrates the corresponding lighting conditions on our system.}
 \label{fig:olat}
\end{figure}

During capture, all the 96 LEDs follow the patterns shown in Fig.~\ref{fig:olat}.

During the capture period, the high-speed camera synchronizes with the 96 LEDs at 1000 fps and outputs an 8-bit Bayer pattern color image stream at a resolution of 2048$\times$1440.

Note that it is required at least 0.1s to acquire a complete dynamic session with 96 LEDs.
Such duration causes misalignment, making it challenging to handle blurriness in the dataset and the low frame rate gives rise to inconsistent dynamic facial capture results.
Inspired by the approach~\cite{meka2020deep}, we interleave ``tracking frames'' into the capture sequence during every six images rather than a complete session to conquer such drawbacks and cast the tracking frames as references, as shown in Fig.~\ref{fig:optical_flow}.
\begin{figure}[t]
  \includegraphics[width=\linewidth]{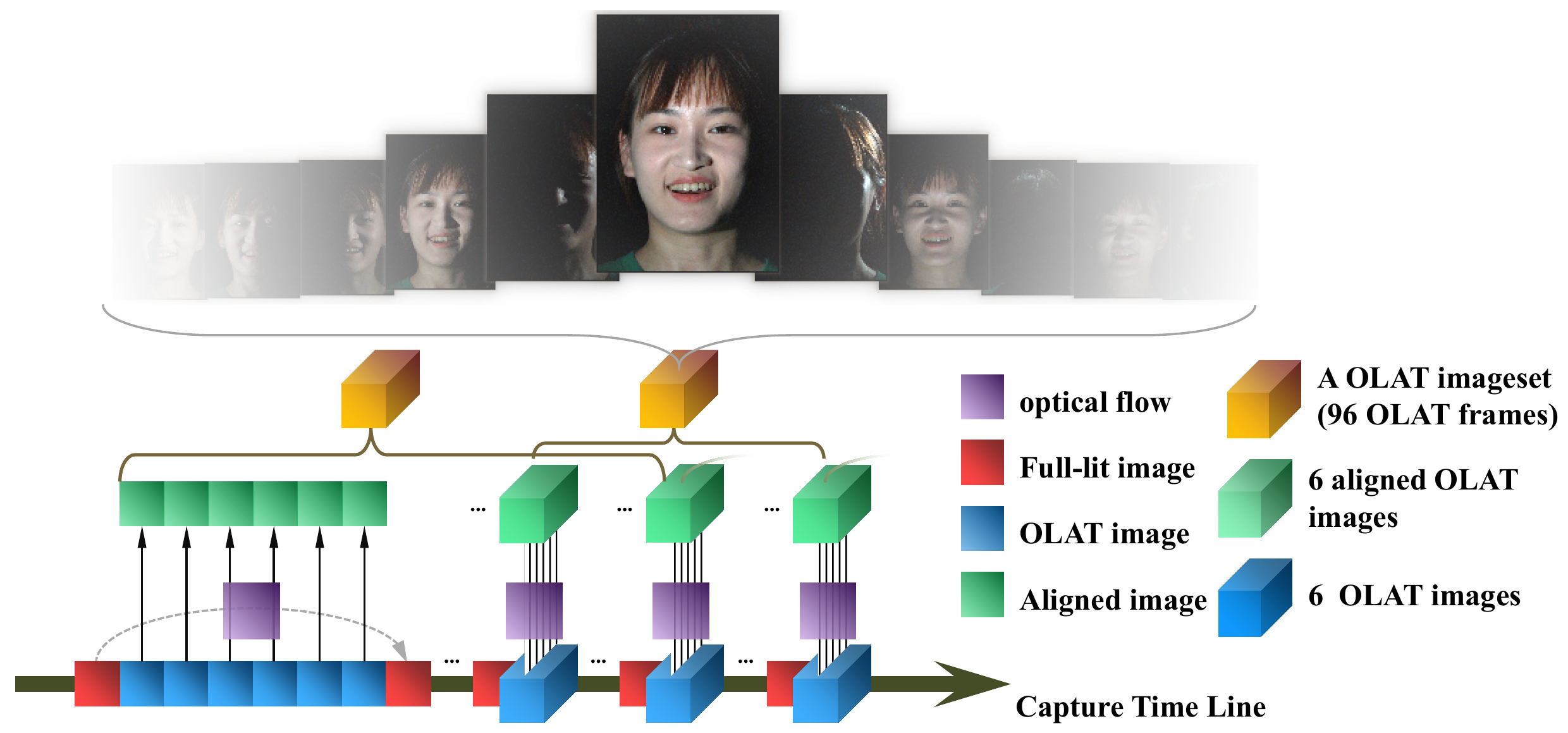}
  \caption{The illustration of capture timing and frame alignment. our method computes optical flows between full-lit images and warps OLAT images accordingly. The "overlapping" strategy allow us to reuse a same OLAT image in different OLAT image set so that we can achieve higher capture frame rate.}
  \label{fig:optical_flow}
\end{figure}

Instead of capturing an image with homogeneous illumination for every 96 images, we capture an image for tracking purposes every 6 images. 
This capture strategy allows us to align the OLAT data between 14 consecutive groups of full-bright frames in any pose to the middle frame with optical flow. 
It is equivalent to that the image between every two homogeneous illumination frames is multiplexed thirteen times to enhance the final optical flow result. 

\section{Normalization Operation Process}
Note that when the source and target actors are different, their facial geometry will be different even in similar facial expressions, which will manifest in facial characteristics, such as eye size, nose length, mouth curvature, etc.
We expect the distribution of conditioning feature maps generated from the source actor is similar to one of the target actor, so as to preserve target-aware appearance rendering results. 

Specifically, we assume parameters as random variables which follow the normal distributions. 
For each parameter, we normalize both the mean and variance of the source actor's distribution $\mathbf{X}_s\sim N(\mathbf{\mu}_s, \mathbf{\sigma}_s^2)$ to be the same as the target actor's distribution $\mathbf{X}_t\sim N(\mathbf{\mu}_t, \mathbf{\sigma}_t^2)$. 
Thus, the normalized parameter $\hat{X}_s$ can be formulated as:
\begin{equation}
\begin{aligned}
\hat{X}_s = (\mathbf{\sigma}_t \oslash \mathbf{\sigma}_s)\circ(\mathbf{X}_s - \mathbf{\mu}_s)+\mathbf{\mu}_t \sim N(\mathbf{\mu}_t, \mathbf{\sigma}_t^2), 
\end{aligned}
\label{eq:normalization}
\end{equation}
where $\oslash$ and $\circ$ are element-wise division and product respectively. 
In our implementation, we regard the head pose $\mathbf{\theta}$ and facial expression $\mathbf{\phi}$ of FLAME parameters as two individual random variables and use the Eqn.~\ref{eq:normalization} for normalization.

Thus, the conditioning feature maps $\mathbf{F}_t$ at time $t$ are formulated as:
\begin{equation}
\begin{aligned}
\mathbf{F}_t = \{\mathbf{\tilde{I}}_t^{d}, \mathbf{\tilde{I}}_t^{c}, \mathbf{\tilde{I}}_t^{l}\},
\end{aligned}
\label{eq:conditioning_feature}
\end{equation}
where $\mathbf{\tilde{I}}_t^{d}, \mathbf{\tilde{I}}_t^{c} = \Pi(G(\mathbf{\beta}, \mathbf{\tilde{\theta}}, \mathbf{\tilde{\phi}}), \mathbf{H})$ and $\Pi$ is the rasterization and texture mapping function; $\mathbf{\tilde{\theta}}$ and $\mathbf{\tilde{\phi}}$ are normalized parameters.

\begin{figure*}[t]
 \includegraphics[width=\linewidth]{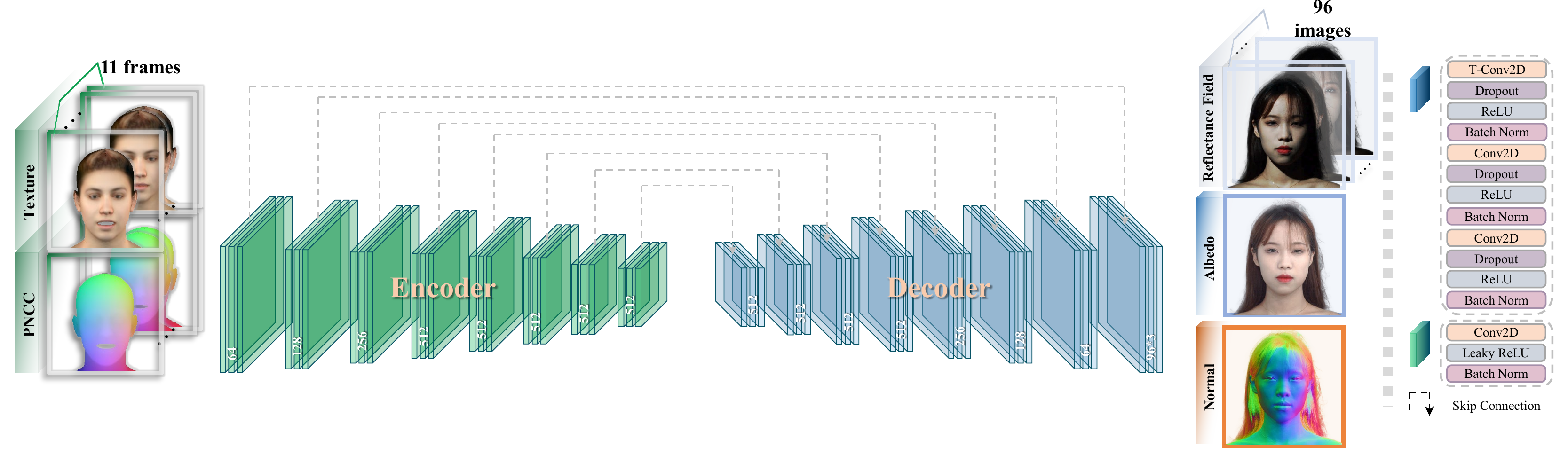}
 \caption{Our multi-frame multi-task architecture design for our rendering-to-video network.}
 \label{fig:network}
\end{figure*}
\section{Network Architecture}

We adopt the U-Net architecture to the proposed translation network as illustrated in Fig.~\ref{fig:network}. 
Our network inferences both the facial reflectance field, the normal, and the albedo simultaneously to facilitate portrait video composition applications.
The proposed network consists of an encoder and a decoder module. 
The encoder extracts multi-scale latent representations of conditioning feature maps, while the decoder module generates the reflectance field, the normal, and the albedo.

Such a multi-task framework enforces the network to learn contextual information of the target actor, so as to produce more detailed reflectance fields. 

\section{Quantitative comparisons}
\begin{figure*}[t]
 \includegraphics[width=\linewidth]{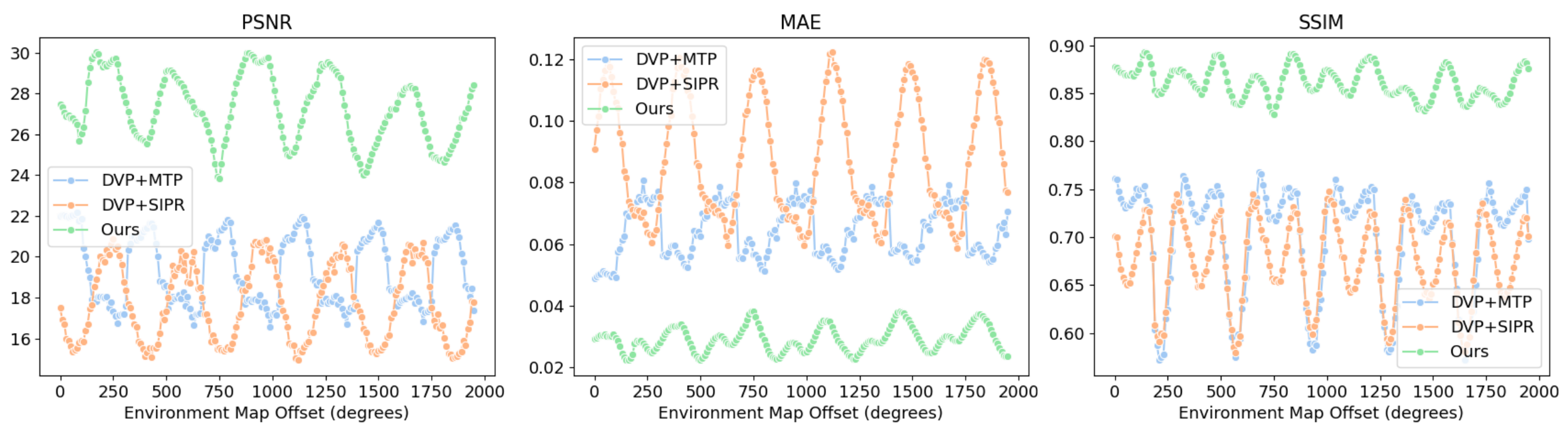}
 \caption{Quantitative comparisons on the test reference data with ground truth. Our results are more realistic and closer to ground truth.}
 \label{fig:loss}
\end{figure*}

For quantitative comparison, we evaluate our rendering results via three various metrics: signal-to-noise ratio (\textbf{PSNR}), structural similarity index(\textbf{SSIM}) and mean absolute error (\textbf{MAE}) to compare with existing state-of-art approaches. 
For comparison, we generate a sequence of relighted portraits from all reference views which are evenly spaced in range of illumination angles.
As shown in Fig.~\ref{fig:loss} due to the high-quality reflectance field inference and fully disentanglement, our method enables entire photo-realistic video portraits synthesis under various illumination conditions, making the rendering quality of any temporal sequence surpassing other baselines with higher PSNR, SSIM and lower MAE.

\begin{figure*}[t]
 \includegraphics[width=\linewidth]{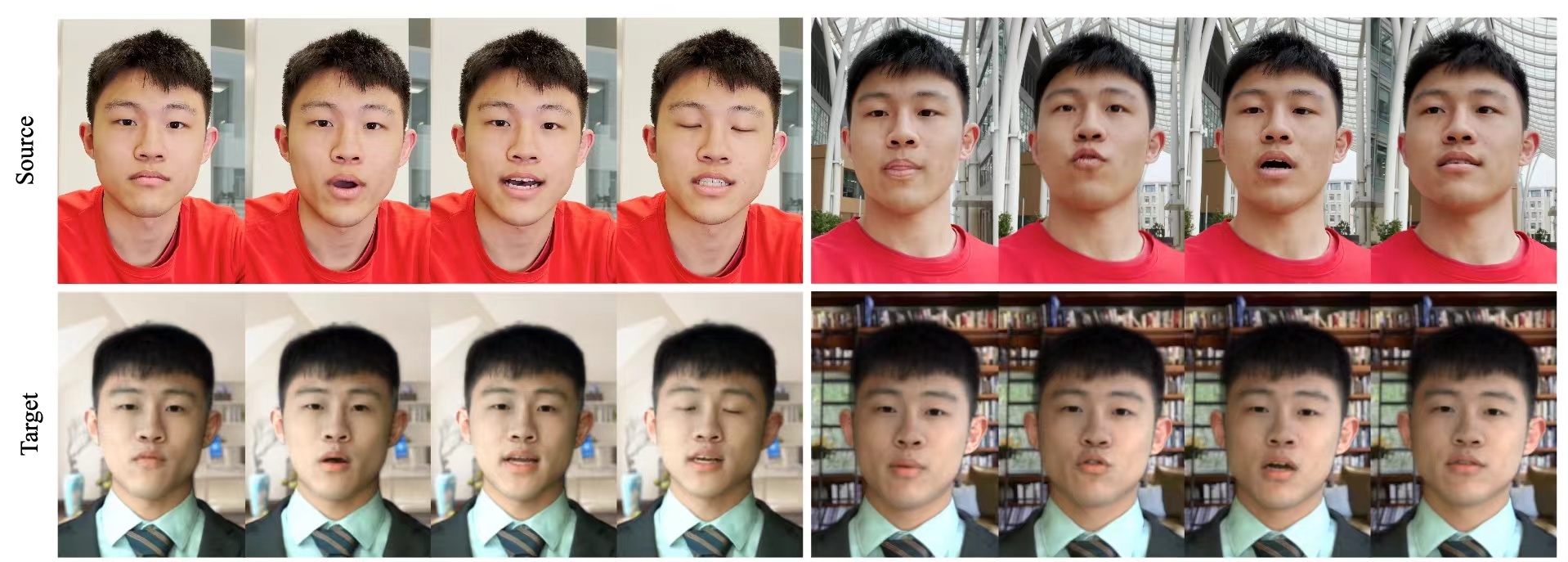}
 \caption{We demonstrated the application of our algorithm in the virtual conference. Users can participate in video conferences on formal occasions in a widerrange of scenarios. Through the control of pose, we can realize that the user still presents a decent participation state in scenes such as walking and notlooking at the camera.}
 \label{fig:vconf}
\end{figure*}

\section{Application: Virtual Conference}
By utilizing our generated dynamic facial reflectance field, we can achieve a relightable virtual conference as shown in Fig.~\ref{fig:vconf}. No matter whether the user is dressing casually with messy hair, or lying on the sofa with a cup of coffee, the user can appear decently in front of others in suits. The conference background can be changed arbitrarily, the most impressive effect is that the light conditions on the generated user in suits will perfectly match the environment. Furthermore, if the user is walking on the street in a rush with a shaking camera, or even he is not looking and the camera, by explicitly controlling the pose parameters, our approach can always let the user look like attending the conference naturally.


\end{document}